\pdfoutput=1
\documentclass{article}

\PassOptionsToPackage{numbers,compress}{natbib}
\usepackage[main, final]{neurips_2026}

\usepackage[utf8]{inputenc}
\usepackage[T1]{fontenc}
\usepackage{hyperref}
\usepackage{url}
\usepackage{booktabs}
\usepackage{amsfonts}
\usepackage{amsmath}
\usepackage{amssymb}
\usepackage{amsthm}
\usepackage{nicefrac}
\usepackage{microtype}
\usepackage{xcolor}
\usepackage{enumitem}
\usepackage{graphicx}

\newcommand{\synNManifolds}{150}
\newcommand{\synNFlat}{13}
\newcommand{\synRateRsq}{0.90}
\newcommand{\synRateSingleD}{0.82}
\newcommand{\synRateSingleCurv}{0.09}
\newcommand{\synFloorRsq}{0.76}
\newcommand{\synFloorSingleCurv}{0.51}
\newcommand{\synFloorSingleD}{0.23}
\newcommand{\synFlatMax}{2\times 10^{-5}}
\newcommand{\synLowCurvLo}{6\times 10^{-4}}
\newcommand{\synLowCurvHi}{4\times 10^{-3}}

\title{The Geometric Wall: Manifold Structure Predicts Layerwise Sparse Autoencoder Scaling Laws}

\author{
  Eslam Zaher$^{1,2}$, Maciej Trzaskowski$^{1,3,4}$, Quan Nguyen$^{1,3,5}$, Fred Roosta$^{1,2}$ \\
  $^1$\,ARC Training Centre for Information Resilience (CIRES) \\
  $^2$\,School of Mathematics and Physics, University of Queensland \\
  $^3$\,Institute for Molecular Bioscience, University of Queensland \\
  $^4$\,Profenso \quad\quad $^5$\,QIMR Berghofer Medical Research Institute
}

\begin{document}

\maketitle

\begin{abstract}

Sparse autoencoders (SAEs) operationalise the linear representation hypothesis: they reconstruct model activations as sparse linear combinations of dictionary atoms hypothesised to represent meaningful linear directions. Their reconstruction error varies sharply across layers in ways that existing scaling laws, fitted at a single layer, do not explain. We contend that this variation is the empirical trace of the changing geometric structure of activation manifolds across depth: as their intrinsic dimension and curvature vary from layer to layer, so does the difficulty of approximating them with sparse linear dictionaries, and SAE width-sparsity scaling becomes a layer-dependent function of manifold structure rather than a single universal law. We conduct the first cross-layer SAE scaling study, fitting per-layer scaling-law surfaces to 844 residual-stream Gemma Scope SAE checkpoints spanning all 68 layers of Gemma 2 2B and 9B, and relating the fitted scaling parameters to four geometric summaries of the activation manifold. We find that manifold geometry predicts the per-layer width-scaling exponent in both models, explaining variation that cannot be reliably captured by layer depth alone. A controlled experiment on 150 synthetic manifolds, where intrinsic dimension and curvature vary independently, further reveals two distinct geometric contributions: intrinsic dimension governs the rate at which reconstruction improves with dictionary width, while curvature governs the residual error at the largest trained width. At the six Gemma showcase layers where richer width grids permit identification of an asymptotic floor, the fitted floors follow the same geometric ordering, consistent with the second-order residual that sparse linear approximations of curved manifolds can leave behind. SAEs thus encounter not merely a finite-resource ceiling but a geometric wall set by the manifold they are trying to reconstruct.
\end{abstract}

\section{Introduction}

SAE reconstruction error varies substantially across the layers of a language model, yet the scaling law that describes SAE reconstruction has been fitted only at a single layer of GPT-4~\citep{gao2024scaling}. Whether that law holds at other layers, and what predicts its variation across depth, has not been measured. This leaves a consequential gap for the interpretability programme that builds on SAEs, whose reconstruction fidelity under a given sparsity constraint matters at every layer we wish to interpret. In particular, it remains unclear whether the reconstruction difficulty observed at different depths reflects differences in dictionary capacity alone or properties of the representations being approximated. Understanding this variation requires looking beyond the scaling behaviour of an individual layer and examining how it relates to the structure of activations throughout the model.

Current practice rests on the linear-representation hypothesis (LRH)~\citep{park2023linear,elhage2022toy}, as adopted by mechanistic interpretability~\citep{cunningham2023sparse,bricken2023monosemanticity,templeton2024scaling,lieberum2024gemma}. SAEs operationalise LRH by encoding each model activation as a sparse combination of dictionary atoms hypothesised to represent meaningful linear directions. The underlying premise is that a sufficiently large and expressive dictionary can represent the structure of activations using few active features at a time. \citet{gao2024scaling} formalised the resulting scaling picture as a joint power law for SAE reconstruction loss $L$ in dictionary width $n$ and sparsity $k$, fitted at a single layer of OpenAI's GPT-4. This relates reconstruction error to the resources available to a sparse linear representation. Its behaviour across depth, however, remains largely unexplored, particularly as the nature of the representations being reconstructed changes from one layer to another.

Parallel results expose tensions in the SAE programme that suggest the LRH-style picture of activations as globally served by a sparse linear basis may be incomplete. A growing literature characterises semantic features in language models as forming non-trivial geometric structures, ranging from circular features for cyclic concepts to hierarchically organised polytopes and curved feature manifolds discretised by sparse families~\citep{engels2024not,li2024geometry,park2025categorical,olah2024multidimensional,michaud2025manifolds,modell2025origins,gurnee2026manifolds}. SAE reconstruction is also uneven across depth. At the middle layers of the same Gemma~2 models we study, the nonlinear residual reconstruction error remains roughly constant across an order of magnitude of dictionary width~\citep{engels2024darkmatter}. Substituting an SAE reconstruction for a model layer's activation can also inflate the model's downstream loss several-fold beyond what a same-magnitude random perturbation would, with the severity varying by layer~\citep{gurnee2024pathological}. These observations concern different aspects of sparse reconstruction, yet together they raise a question that current scaling analyses do not address: what determines why some layers are substantially harder to reconstruct than others, even as dictionary capacity increases?

On the geometric side, a research community studying activation manifolds has independently mapped their depthwise structure. Intrinsic dimension follows a \emph{hunchback} profile that peaks at middle layers across architectures~\citep{ansuini2019intrinsic,valeriani2023geometry}; intrinsic dimension and curvature both vary systematically with depth in language models~\citep{mabrok2026latent}; and information-geometric analyses identify a non-Euclidean structure on activation space through the pullback of the Fisher--Rao metric from the model's predictive distribution~\citep{park2026infogeom,mabrok2026latent}. These results establish that the geometry of activation representations changes considerably across depth, with the middle layers often exhibiting particularly high intrinsic dimension. Strikingly, this is also the region where SAE reconstruction has been observed to improve little despite substantial increases in dictionary width. The two literatures thus describe depth-dependent phenomena that may be closely related: one documents variation in the difficulty of sparse reconstruction, while the other characterises variation in the geometric complexity of the representations being reconstructed. Yet these lines of research have developed largely independently, and whether the geometric structure of activations can account for the layerwise behaviour of SAE scaling remains unexamined.

We contend that these depthwise patterns are manifestations of a common underlying phenomenon: the geometric structure of activation representations conditions how well they can be approximated by sparse linear dictionaries. The variation in SAE reconstruction difficulty across layers may therefore reflect the same geometric changes that have been independently documented in the activation-manifold literature. In this account, the scaling law may retain its functional form across layers, while its parameters reflect the changing geometric structure of the activation manifolds being approximated. In particular, both the rate at which reconstruction error decreases with dictionary width and the residual loss approached at large widths may depend on the intrinsic dimension and curvature of the underlying representations. We examine this connection through four geometric summaries of activation space: intrinsic dimension, multi-scale curvature, tangent variation, and within-layer heterogeneity. The central question is whether these quantities account for systematic variation in SAE scaling beyond what can be explained by layer depth alone. Such a relationship would place the observed limits of sparse reconstruction within a broader account of the geometric structure of language-model representations.

We investigate this relationship using the publicly released Gemma Scope SAEs~\citep{lieberum2024gemma}, comprising all $844$ JumpReLU checkpoints spanning the $26$ layers of Gemma~2 2B and the $42$ layers of Gemma~2 9B. We fit the reconstruction scaling law separately at each layer and examine how its parameters vary with the corresponding activation geometry. We compare the explanatory power of the geometric descriptors against depth-only baselines under cross-validation that accounts for dependencies between neighbouring layers. Finally, we train SAEs on synthetic manifolds whose intrinsic dimension and curvature vary independently, allowing their contributions to reconstruction scaling to be examined without the depth-dependent correlations present in language models.

\paragraph{Contributions.} Our contributions are threefold.

\textbf{(i) Geometric framework for the SAE residual.} A pullback information-geometric account of why activation spaces need not admit globally efficient sparse linear codes, yielding a directional prediction tested below: layers with higher intrinsic dimension and curvature should scale less efficiently in dictionary width.

\textbf{(ii) Geometry-conditioned scaling law.} We extend \citet{gao2024scaling}'s width-sparsity scaling law to a geometry-conditioned account in which the scaling-law parameters become layerwise functions of activation manifold geometry, fit per model and evaluated against depth baselines (full method in \S\ref{sec:framework} and \S\ref{sec:experiments}; validation in Appendices~\ref{app:validation-suite} and~\ref{app:depth}).

\textbf{(iii) Separation of the geometric channels and floor evidence.} (a) Geometry predicts held-out layers and blocks of contiguous layers that depth-only models do not, in both Gemma~2 models; (b) in a controlled experiment on synthetic manifolds with independently varied intrinsic dimension and curvature, intrinsic dimension governs the width-scaling rate and curvature governs the reconstruction error at the largest trained width, a finite-width proxy for the floor; (c) at the Gemma showcase layers where the asymptotic floor is identifiable, the fitted floor follows the predicted ordering in intrinsic dimension and multi-scale curvature, consistent with and extending \citet{gao2024scaling}'s \emph{spectrum of structure} picture of the irreducible loss.

\section{Background}

We fix notation for transformers, sparse autoencoders, and the pullback information geometry used throughout; experiments (\S\ref{sec:experiments}) specialise to the residual stream.

\subsection{Language models and hidden representations}

A transformer with $\mathcal{L}$ layers maps each hidden state $h_\ell \in \mathbb{R}^{d_\ell}$ to a next-token distribution via the \emph{predictive map}
\begin{equation}\label{eq:predictive-map}
    F_\ell: \mathbb{R}^{d_\ell} \to \Delta^{V-1}_{\circ}, \qquad F_\ell(h) = \mathrm{softmax}(z_\ell(h)),
\end{equation}
where $z_\ell$ composes layers $\ell$ through unembedding. For a data distribution $\mathcal{D}$, the activation set $\mathcal{H}_\ell \subseteq \mathbb{R}^{d_\ell}$ is the object that SAEs approximate.

\subsection{Sparse autoencoders as sparse linear approximation}

A sparse autoencoder at layer $\ell$ encodes $h \in \mathbb{R}^{d_\ell}$ as a sparse code $a = \phi(W^{\mathrm{enc}} h + b^{\mathrm{enc}}) \in \mathbb{R}^n$ with $n \gg d_\ell$ and decodes $\hat{h} = W^{\mathrm{dec}} a + b^{\mathrm{dec}}$, the columns of $W^{\mathrm{dec}}$ being the dictionary atoms. Principal variants (ReLU~\citep{cunningham2023sparse,bricken2023monosemanticity}, TopK~\citep{gao2024scaling}, and JumpReLU~\citep{rajamanoharan2024jumping,lieberum2024gemma}) differ in $\phi$ but share the sparse-linear structure. Reconstruction quality is measured by the per-sample normalised mean-squared error $\mathrm{NMSE}(h_i, \hat{h}_i) = \|h_i - \hat{h}_i\|^2 / \|h_i\|^2$, aggregated to a checkpoint-level loss $L$ (Section~\ref{sec:error-metrics}).

\subsection{The SAE scaling law}

\citet{gao2024scaling} showed that reconstruction loss follows a joint scaling law in dictionary width $n$ and realised sparsity $k = L_0$, fitted at a layer $5/6$ of the way into a GPT-4-series model on the post-layernorm residual stream activations. They first established a no-floor power-law surface $L(n,k) = A(k)\,n^{-\alpha(k)}$, then introduced the asymptotic floor $B(k)$ as a refinement at the same layer, where richer width coverage exposed the saturation, yielding the with-floor form
\begin{equation}\label{eq:scaling-law}
    L(n, k) = A(k)\, n^{-\alpha(k)} + B(k),
\end{equation}
with $A(k) = \exp(a_0 + \beta_k \log k)$, $\alpha(k) = -(\beta_n + \gamma \log k)$, $B(k) = \exp(\zeta + \eta \log k)$. Here $A(k)\,n^{-\alpha(k)}$ is the \emph{reducible loss}, whose width exponent $\alpha(k)$ depends on sparsity through the interaction parameter $\gamma < 0$: wider dictionaries help more when more latents are active. The term $B(k)$ is the \emph{irreducible} floor: variance no dictionary can capture at a given $L_0$. The law was established at a single layer; whether its layerwise behaviour varies systematically with cross-layer structure was not addressed, and is the central empirical question of this paper. The no-floor and with-floor forms are linked through the log-linearisation
\begin{equation}\label{eq:scaling-loglinear}
    \log L \;\approx\; a_0 + \beta_n \log n + \beta_k \log k + \gamma\, \log n \cdot \log k,
\end{equation}
which holds when $B$ is small relative to $A\,n^{-\alpha}$ at the observed widths.

\subsection{Pullback information geometry of activation spaces}\label{sec:geometry}

We develop the geometric framework connecting the model's predictive structure to its hidden representations; full differential-geometric background is in Appendix~\ref{app:geometry}. The natural Riemannian metric on a space of probability distributions is the \textbf{Fisher information metric}~\citep{amari2016information}, unique up to scaling among metrics invariant under sufficient statistics~\citep{chentsov1982statistical}. For categorical distributions $\boldsymbol{\pi} \in \Delta^{V-1}_\circ$, the Fisher metric in logit coordinates takes the form of the categorical covariance
\begin{equation}\label{eq:sigma-p}
    \boldsymbol{\Sigma}_{\boldsymbol{\pi}} = \mathrm{diag}(\boldsymbol{\pi}) - \boldsymbol{\pi}\boldsymbol{\pi}^\top,
\end{equation}
of rank $V{-}1$; the resulting manifold $(\Delta^{V-1}_\circ, \mathbf{G}_{\mathrm{FR}})$ has Bhattacharyya arc length as its geodesic distance~\citep{bhattacharyya1943measure,amari2016information}.

\paragraph{Pullback to hidden layers.}
The predictive map $F_\ell$ (Eq.~\ref{eq:predictive-map}) pulls the Fisher--Rao metric back to $\mathbb{R}^{d_\ell}$: with $J_\ell(h) = \partial z_\ell / \partial h$,
\begin{equation}\label{eq:pullback-fisher}
    \mathbf{G}_\ell(h) = J_\ell(h)^\top\; \boldsymbol{\Sigma}_{F_\ell(h)}\; J_\ell(h).
\end{equation}
The rank of $\mathbf{G}_\ell(h)$ is at most $\min(V{-}1, \mathrm{rank}\,J_\ell(h))$; directions in $\ker J_\ell$ are perturbations invisible to the output. Because $\mathbf{G}_\ell$ depends on the local hidden state through both the Jacobian and the output distribution, activations close in Euclidean distance may be far apart under the pullback metric, and vice versa (Appendix~\ref{app:geometry}).

\paragraph{Two roles of geometry, and the geometric mismatch.}

Activations $h_\ell \in \mathbb{R}^{d_\ell}$ encode information about the next-token distribution $F_\ell(h)$ they induce, providing a natural functional geometry through the pullback Fisher--Rao metric $\mathbf{G}_\ell$ (Eq.~\ref{eq:pullback-fisher}). Under this metric, distances reflect changes in the model's predictive distribution, and its degeneracy identifies directions in activation space that are locally indistinguishable in their predictive effects. This provides a functional perspective on the low-dimensional, curved structures that activation representations may exhibit. Standard SAEs, however, are trained to minimise the ambient $\ell_2$ reconstruction loss on activation vectors~\citep{cunningham2023sparse,bricken2023monosemanticity,gao2024scaling,lieberum2024gemma}, which measures reconstruction fidelity under the Euclidean metric rather than the pullback metric. This distinction exposes two geometric limitations: a \emph{basis--manifold} mismatch, where a flat $k$-sparse linear dictionary must approximate activations distributed along a curved manifold in the ambient space; and an \emph{objective--metric} mismatch, where minimising Euclidean reconstruction error need not preserve distances measured by the model's predictive geometry. Our empirical analysis focuses on the former, using the extrinsic Euclidean descriptors $\{d_{\mathrm{int}}, \kappa_{\mathrm{ms}}, \kappa^{\mathrm{tv}}, \nu\}$ to characterise the activation geometry that SAE dictionaries are trained to approximate (Appendix~\ref{app:trimming}). The pullback Fisher--Rao geometry motivates the broader geometric perspective but is not computed or used in our experiments. We discuss the relationship between the two metrics and the conditions under which their geometric properties carry over in Appendix~\ref{app:geometry}. This connection also points towards a more functionally grounded formulation of sparse feature learning, in which the geometry of the model's predictive distribution informs the construction of features and their reconstruction objective, which we leave for future work.

\paragraph{Geometric summaries.}
We characterise $\mathcal{H}_\ell$ through four layerwise scalar summaries that quantify aspects of local manifold structure:
\begin{itemize}[nosep,leftmargin=1.5em]
    \item \emph{Intrinsic dimension} $d_{\mathrm{int},\ell}$: the number of local degrees of freedom of the manifold.
    \item \emph{Multi-scale curvature} $\kappa_{\mathrm{ms},\ell}$: the average departure of the manifold from its tangent plane across local neighbourhoods of varying radius.
    \item \emph{Tangent variation} $\kappa^{\mathrm{tv}}_\ell$: the rate at which the tangent space rotates between nearby points.
    \item \emph{Heterogeneity} $\nu_\ell$: the average within-neighbourhood variation in local intrinsic dimension across the layer, capturing how non-uniform the manifold's intrinsic dimension is at small scales.
\end{itemize}
Each controls sparse linear approximation efficiency: higher intrinsic dimension demands more atoms to span the tangent structure; higher curvature shrinks each atom's region by introducing second-order departure from any tangent approximation; higher heterogeneity prevents any single dictionary geometry from being uniformly efficient across the manifold. Estimation procedures are detailed in Appendix~\ref{app:trimming} (two-nearest-neighbour (TWO-NN)~\citep{facco2017estimating} for $d_{\mathrm{int}}$, multi-scale principal-component-analysis (PCA) residual estimator for $\kappa_{\mathrm{ms}}$, neighbour Gauss-map estimator for $\kappa^{\mathrm{tv}}$, pointwise-dimension dispersion estimator for $\nu$).

\section{Geometry-Conditioned Account of SAE Scaling}\label{sec:framework}

We ask whether the layerwise behaviour of the surface $L_\ell(n, k)$ is conditioned by activation geometry $\mathbf{g}_\ell$. Two structural constraints from the released SAE family shape our methodology. First, only two \emph{backbone} dictionary widths are shared across all $68$ layers ($n_{\mathrm{lo}} = 16$K, $n_{\mathrm{hi}} = 65$K for 2B; $n_{\mathrm{lo}} = 16$K, $n_{\mathrm{hi}} = 128$K for 9B); a $6$-parameter with-floor surface (Eq.~\ref{eq:scaling-law}) is therefore not identifiable per-layer at most layers. Second, \citet{gao2024scaling} themselves first fit a no-floor power law to their SAE data, then introduced the asymptotic floor $B(k)$ as a refinement at the layer where richer width coverage made the floor identifiable. We follow the same natural progression: a per-layer no-floor surface fit at \emph{every} layer (\S\ref{sec:two-stage}), and a refit with the with-floor form at the six \emph{showcase} layers per model where the grid contains $\geq 3$ widths (\S\ref{sec:showcase-calibration}). The cross-layer geometric question is then answered by a Stage~$2$ regression of the per-layer Stage~$1$ scaling-law parameters and derived width exponent against the layerwise geometric summaries (\S\ref{sec:stage2}).

\subsection{Stage 1: per-layer no-floor surface fits}\label{sec:two-stage}

At each layer $\ell$ we fit the floor-free log-linearisation of Eq.~\ref{eq:scaling-loglinear} as a $4$-parameter surface,
\begin{equation}\label{eq:per-layer-nofloor}
    \log L_\ell(n, k) \;=\; a_{0,\ell} + \beta_{n,\ell}\,\log n + \beta_{k,\ell}\,\log k + \gamma_\ell\,\log n\,\log k,
\end{equation}
on the layer's full grid of $(n, k, L)$ triples (interpolated across the layer's realised sparsities by piecewise cubic Hermite interpolating polynomial (PCHIP); Section~\ref{sec:experiments}). The fit is identifiable from the two backbone widths shared across all layers and tightens at the six showcase layers with $\geq 3$ widths. From the per-layer parameters $(\beta_{n,\ell}, \gamma_\ell)$ we read off the layerwise width-scaling exponent at any sparsity target $k$,
\begin{equation}\label{eq:alpha-derived}
    \alpha_\ell(k) \;=\; -\bigl(\beta_{n,\ell} + \gamma_\ell\,\log k\bigr).
\end{equation}

Because the fit assumes $B = 0$, we refer to $\alpha_\ell(k)$ from Eq.~\ref{eq:alpha-derived} as the \emph{effective no-floor exponent} over the observed width range. It coincides with the asymptotic exponent only when the floor's contribution at the observed widths is small; when a non-zero floor is present it underestimates the asymptotic with-floor exponent (Appendix~\ref{app:no-floor-vs-floor}). We show in Appendix~\ref{app:secant} that this no-floor surface parameterisation agrees with the model-free two-width secant estimate. Throughout \S\ref{sec:results} we use $\alpha_\ell(k)$ as the per-layer Stage~$1$ exponent at $k = 50$ unless otherwise stated; the calibration of this exponent against the with-floor fit at showcase layers is reported in \S\ref{sec:showcase-calibration}.

\subsection{Stage 2: cross-layer geometric regression}\label{sec:stage2}

The Stage~$1$ outputs supply three per-layer scalar targets: the derived width exponent $\alpha_\ell(k)$ (Eq.~\ref{eq:alpha-derived}), the width-scaling coefficient $\beta_{n,\ell}$, and the width-sparsity interaction coefficient $\gamma_\ell$. The four layerwise geometric summaries $\mathbf{g}_\ell = (d_{\mathrm{int},\ell}, \kappa_{\mathrm{ms},\ell}, \kappa^{\mathrm{tv}}_\ell, \nu_\ell)$ defined in \S\ref{sec:geometry} enter the regression after a uniform $\log$ transform with light $1/99$-percentile trimming and per-feature standardisation (Appendix~\ref{app:trimming}); tables and figures use raw symbols throughout. We regress each target $y_\ell$ on the layerwise geometric summaries via ordinary least squares (OLS):
\begin{equation}\label{eq:stage2}
    y_\ell \;=\; \mu \;+\; \sum_{p=1}^{P} \theta_p\, g_{\ell,p} \;+\; \varepsilon_\ell, \qquad P \in \{1,2,4\}.
\end{equation}
Regression is fit per model (no across-model pooling), so that model-specific intercepts are not conflated with shared geometric structure.

\paragraph{Hypothesis hierarchy.} For each target $y_\ell$ we compare a nested family of models, each corresponding to a substantive claim about whether and how layerwise geometry shapes the target:
\begin{itemize}[nosep,leftmargin=1.5em]
    \item \textbf{H0} (\emph{geometry-invariant null}): the per-layer target is a constant across layers, $y_\ell = \mu + \varepsilon_\ell$ ($1$ parameter). For $y_\ell = \alpha_\ell(k=50)$, this is the hypothesis that the per-layer width-scaling exponent is the same at every layer, i.e., the scaling law is universal across depth and geometry plays no role. H0 is the natural null because it is what a single-layer scaling-law fit (the form used by \citet{gao2024scaling}) implicitly assumes when extrapolated to other layers.
    \item \textbf{H1$_g$} (\emph{single-feature dependence}): the per-layer target depends on one geometric feature $g$ (2 parameters; one model per feature). Tests whether any single geometric summary explains layerwise variation.
    \item \textbf{H2$_{\mathrm{low}\rho}$} (\emph{lowest-pairwise-correlation feature pair}): the per-layer target depends on the two geometric features with the smallest within-model $|\rho|$ (3 parameters; $d_{\mathrm{int}}+\kappa_{\mathrm{ms}}$ at 9B and $d_{\mathrm{int}}+\nu$ at 2B). Tests whether two minimally-collinear features carry independent additive variance. The complementary best-single + next-best additivity check is the H1$\to$H2$_{\mathrm{best+next}}$ step of the F-test ladder in Appendix~\ref{app:validation-suite}.
    \item \textbf{H$_{\mathrm{full}}$} (\emph{full geometric dependence}): the per-layer target depends on all four features (5 parameters). Tests whether the joint geometric specification adds further explanatory power beyond any pair.
\end{itemize}
Rejection of H0 (in favour of any H1, H2$_{\mathrm{low}\rho}$, or H$_{\mathrm{full}}$) is rejection of the geometry-invariant scaling-law assumption: it shows that the target varies systematically with measurable geometric properties of the activation manifold rather than being a single layerwise constant.

\paragraph{Where geometry enters: intercept vs slope.} The Stage~$1$ exponent $\alpha_\ell(k) = -(\beta_{n,\ell} + \gamma_\ell \log k)$ has two channels through which geometry shapes the curve: the intercept $\beta_{n,\ell}$, which sets the level of the curve, and the interaction $\gamma_\ell$, which sets its tilt with respect to $\log k$. Regressing $\beta_{n,\ell}$ and $\gamma_\ell$ on geometry (Eq.~\ref{eq:stage2}) tests where in the curve the geometric signal sits.

\paragraph{Evaluation.} For each fitted model we report in-sample $R^2$, leave-one-/two-/three-layer-out $R^2$ (LOO/L2O/L3O), the Akaike and Bayesian information criteria (AIC, BIC), and an F-test against H0. Because neighbouring layers are autocorrelated, we also evaluate using contiguous-block cross-validation and a circular-shift null, and compare against depth-only baselines (\S\ref{sec:experiments}, Appendix~\ref{app:depth}).

\subsection{Floor calibration at showcase layers}\label{sec:showcase-calibration}

At the six layers where the released grid supplies $\geq 3$ widths after monotone filtering (layers $\{5, 12, 19\}$ for 2B, and $\{9, 20, 31\}$ for 9B), we re-fit the full $6$-parameter with-floor surface of Eq.~\ref{eq:scaling-law} by joint nonlinear least squares on $\log L$ (procedure, monotone filters, and more technical details are in Appendix~\ref{app:no-floor-vs-floor}). The resulting per-layer asymptotic floor $B_\ell(k)$ estimates the reconstruction error that added dictionary atoms cannot remove at sparsity $k$. Because $B$ is identifiable at only six layers, we report its ordering against the layerwise geometric summaries as suggestive evidence (\S\ref{sec:results}). The floor--geometry relationship is tested systematically in the controlled experiment (\S\ref{sec:experiments}, Appendix~\ref{app:synthetic}), and the curvature mechanism is discussed in Appendix~\ref{app:floor}.

\section{Experimental Setup}\label{sec:experiments}

\paragraph{Models and SAE family.}
We study the residual stream of Gemma~2 2B ($\mathcal{L}{=}26$ layers, $d{=}2304$) and 9B ($\mathcal{L}{=}42$ layers, $d{=}3584$)~\citep{team2024gemma}, using all publicly released JumpReLU SAEs from Gemma Scope~\citep{lieberum2024gemma}. The released checkpoints contain two backbone widths shared across all layers ($16$K and $65$K for 2B; $16$K and $128$K for 9B), and richer sweeps of up to seven widths at three \emph{showcase} layers per model: $\{5, 12, 19\}$ for 2B and $\{9, 20, 31\}$ for 9B. Total: $312$ checkpoints for 2B and $532$ for 9B (844 total). At each (layer, width) bundle, the released SAEs span a range of training sparsity targets, yielding multiple checkpoints at varying $L_0$. Because training budgets vary with dictionary width, our scaling results characterise the released SAE family rather than convergence-matched scaling (Appendix~\ref{app:training-budget}). Full source code for our experiments is released at \url{https://github.com/eszaher/The-Geometric-Wall-in-SAEs}.

\paragraph{Disjoint geometry/error data.} Geometry estimation and error measurement use \emph{disjoint} data, so geometric summaries cannot overfit to token-level artefacts. We partition the Colossal Clean Crawled Corpus (C4) validation set~\citep{raffel2020exploring} by document index (sequences $0$--$5$K for geometry, $5$K--$10$K for error), extracting 50K activation vectors per layer per partition. Cross-corpus validation uses WikiText-103~\citep{merity2017pointer} for the error measurement while keeping geometry on C4.

\paragraph{Reconstruction error and sparsity.}\label{sec:error-metrics}
We observed that a small number of unusually high-norm activations in each layer can disproportionately influence aggregate statistics, obscuring the behaviour of the majority of tokens. To mitigate this effect, we compute the checkpoint-level loss $L$ and sparsity $L_0$ as symmetric trimmed means of the per-token NMSE and active-latent count, respectively, over the same subset of tokens. This ensures that each $(L, L_0)$ pair is computed from a shared token subset by construction (see Appendix~\ref{app:trimming} for preprocessing details, a norm-free cross-check metric, and sensitivity analyses).

\paragraph{Statistical protocol.}
We follow the two-stage estimation procedure introduced in \S\ref{sec:framework}. In Stage~$1$, we estimate the scaling parameters using ordinary least squares (OLS), with piecewise cubic Hermite interpolation (PCHIP)~\citep{fritsch1980monotone} used to estimate reconstruction losses at common sparsity levels across checkpoints. In Stage~$2$, we assess the stability of these estimates by leaving out groups of $K$ layers, for $K\in\{1,2,3\}$. Rather than refitting the model for every held-out group, we compute the leave-$K$-layer-out estimates in closed form using the OLS hat matrix~\citep{allen1974relationship,stone1974cross}. For the showcase layers, we also fit the model with a nonzero loss floor, jointly optimising its six parameters using multi-start Nelder–Mead~\citep{nelder1965simplex}. A comparison of the models with and without a loss floor is provided in Appendix~\ref{app:no-floor-vs-floor}.

We separately evaluate whether the geometric predictors explain variation across layers beyond what can be accounted for by layer depth alone. For this comparison, we use the model-free secant exponent $\alpha^{\mathrm{eff}}_\ell(k=50)$, which is available for $41$ of the $42$ layers in the 9B model. Predictive performance is evaluated using four-fold contiguous-block cross-validation, in which adjacent layers are held out together to account for their dependence. We additionally consider a purged variant that excludes the immediate neighbouring layer on either side of each held-out block from the training set, reducing the influence of local correlations between training and test layers. All predictors are evaluated on identical folds, with no model selection performed separately within each fold. To assess statistical significance while preserving the autocorrelation structure across layers, we compare the observed associations against a circular-shift null distribution. The full evaluation protocol is described in Appendix~\ref{app:depth}.

Additional validation experiments, including nested-model F-test comparisons, cross-model transfer, and leave-one-$L_0$-out cross-validation of the PCHIP interpolation, are reported in Appendix~\ref{app:validation-suite}.

\paragraph{Controlled synthetic experiment.}
To isolate the effects of geometry from those of layer depth, we construct \synNManifolds{} synthetic manifolds with independently sampled intrinsic dimension $d \in [4,16]$ and curvature amplitude $a \in [0.3,5.0]$, including \synNFlat{} exactly flat controls (Eq.~\ref{eq:synthetic-manifold}; Appendix~\ref{app:synthetic}). All manifolds share the ambient dimension $D=128$. We train grouped TopK SAEs with fixed sparsity $k=32$ at five dictionary widths ranging from $256$ to $4096$ ($2\times$ to $32\times$ overcomplete), using the same optimiser and convergence criterion throughout. By construction, intrinsic dimension and curvature vary independently, while every manifold is evaluated over the same range of dictionary widths under a uniform training protocol. This allows us to examine their contributions to SAE scaling without the confounding effects of layer depth or width-dependent training budgets.

\section{Results}\label{sec:results}

\begin{figure}[t]
    \centering
    \includegraphics[width=0.95\textwidth]{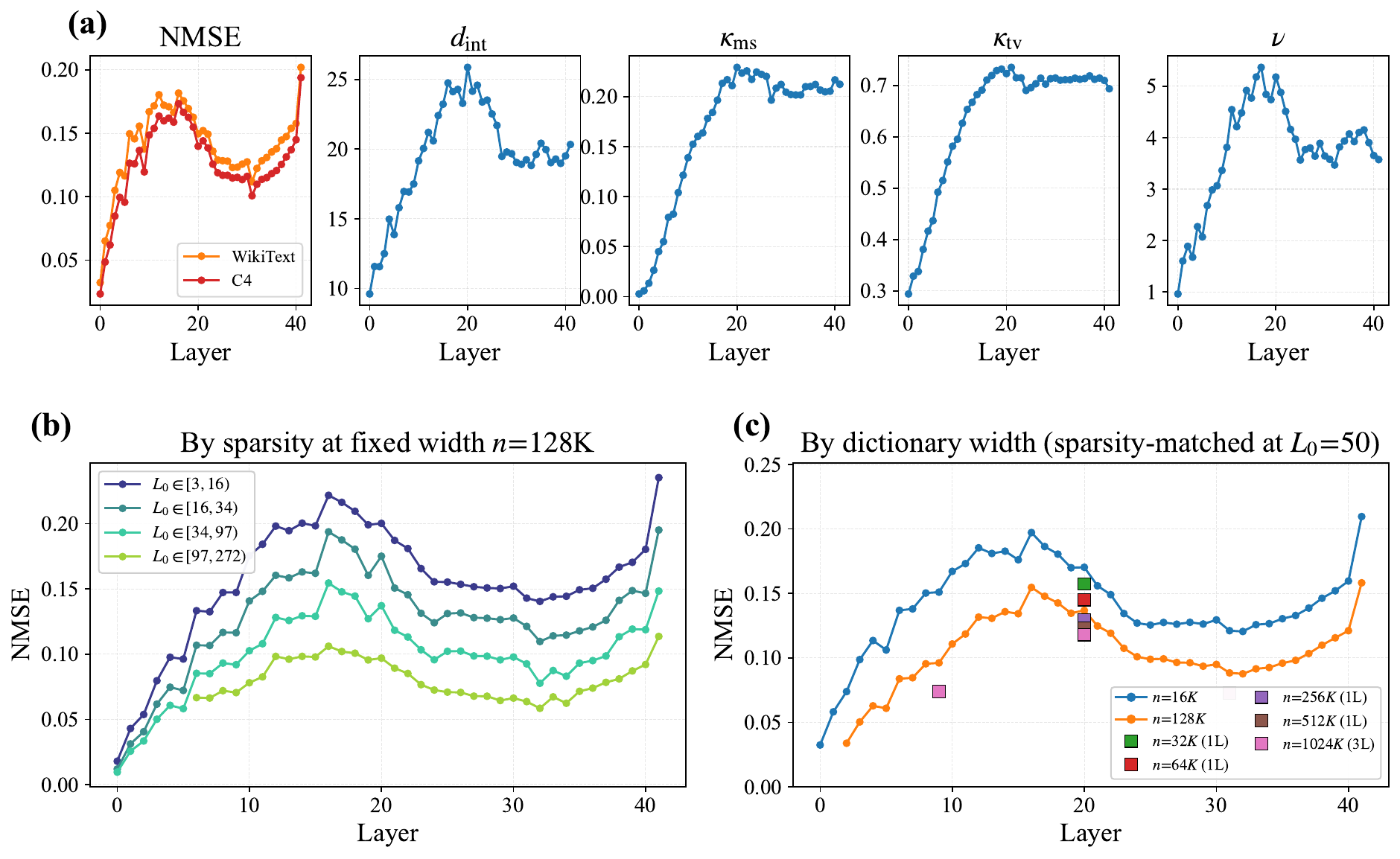}
    \caption{Phenomenology of layerwise SAE reconstruction for Gemma~2 9B. \textbf{(a)}~Layerwise checkpoint-level NMSE profile (C4 and WikiText-103) alongside the four geometric summaries; NMSE peaks mid-depth and the two corpora track each other closely. \textbf{(b)}~NMSE at the upper backbone width ($n = 128$K) stratified by $L_0$ bin; increasing $L_0$ shifts the level without reshaping the depth profile. \textbf{(c)}~NMSE per layer per dictionary width, sparsity-matched at $L_0 = 50$ via PCHIP; the two backbone widths ($16$K, $128$K) cover all 42 layers, while $32$K, $64$K, $256$K, $512$K appear at the central showcase layer and $1024$K at all three showcase layers. 2B figure: Fig.~\ref{fig:profiles-2b}.}
    \label{fig:profiles}
\end{figure}

\paragraph{Geometry and reconstruction co-vary across depth.} Figure~\ref{fig:profiles} presents the phenomenological views of the released SAE family that motivate the analysis. Panel~(a) shows the layerwise checkpoint-level NMSE profile (on C4 and WikiText-103) alongside the four layerwise geometric summaries. The NMSE profile rises through early layers, peaks at mid-depth, dips through late-mid layers, and exhibits a sharp uptick at the final layers; C4 and WikiText curves track each other throughout. The geometric summaries show partly overlapping but distinct depth signatures: intrinsic dimension and heterogeneity peak at mid-depth and decline toward the end, while curvature and tangent variation rise through early layers and plateau without returning. Panel~(b) stratifies the layerwise NMSE at the upper backbone width by sparsity bin ($L_0$); increasing $L_0$ uniformly lowers the error level, but the depth profile is preserved across every sparsity stratum, so sparsity shifts the absolute level without reshaping the depth profile. Panel~(c) stratifies the layerwise NMSE (averaged over $L_0$) by dictionary width. The two backbone curves run roughly parallel in absolute terms across most of the network, but the multiplicative factor between them shrinks with depth, indicating that per-layer width-scaling efficiency varies with depth, which is the quantity we now characterise via the two-stage fit of \S\ref{sec:two-stage}.

\begin{figure}[!htbp]
    \centering
    \includegraphics[width=\textwidth]{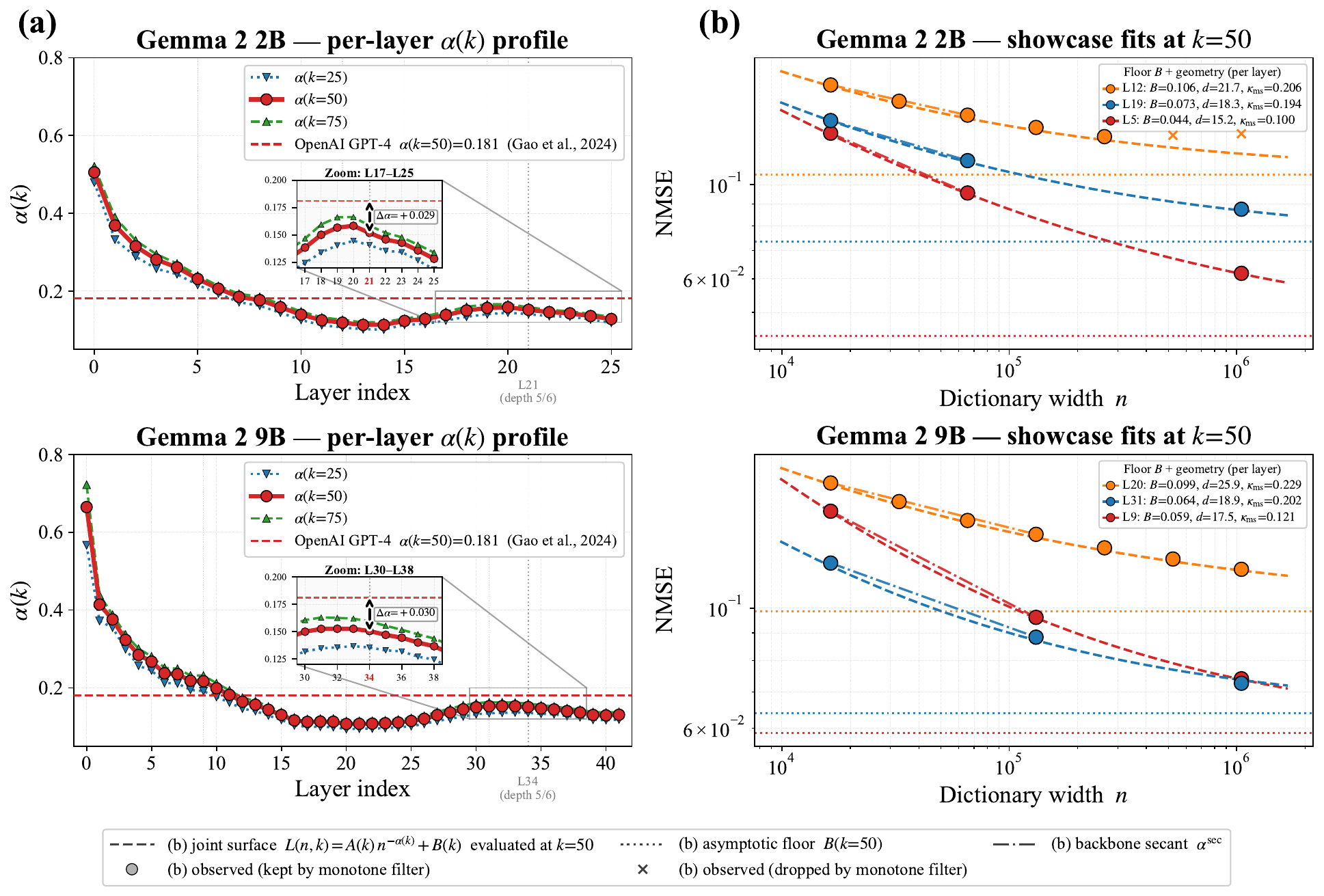}
    \caption{Per-layer width-scaling exponents and the geometric wall, in a $2\times2$ grid (rows: models 2B/9B; columns: panels). \textbf{(a) Per-layer $\alpha_\ell(k)$ profiles.} Stage~$1$ per-layer $\alpha_\ell(k)$ from Eq.~\ref{eq:alpha-derived} at $k \in \{25, 50, 75\}$ (blue dotted, red solid, green dashed; $k = 50$ thicker). The dashed horizontal line marks the OpenAI value $\alpha(k=50) = 0.181$ at relative depth $5/6$ of GPT-4~\citep{gao2024scaling}; a vertical grey dotted line marks the depth-$5/6$ layer. Zoomed insets (one per row) annotate the $\sim 0.03$ gap $\Delta\alpha$ between our $\alpha_\ell(k=50)$ at depth $5/6$ and Gao's $0.181$, consistent with the no-floor underestimation analysis (Appendix~\ref{app:no-floor-vs-floor}). \textbf{(b) Showcase-layer with-floor surface fits at $k = 50$.} Joint $6$-parameter fits (Eq.~\ref{eq:scaling-law}, Appendix~\ref{app:no-floor-vs-floor}) at the three showcase layers per model (2B $\ell \in \{5, 12, 19\}$, 9B $\ell \in \{9, 20, 31\}$); dotted horizontal lines mark the fitted floor $B_\ell(k = 50)$, whose vertical ordering tracks the layers' $d_{\mathrm{int}}$ and $\kappa_{\mathrm{ms}}$ ordering in both models.}
    \label{fig:exponents}
\end{figure}

\paragraph{Per-layer width-scaling exponents are predicted by manifold geometry.} Figure~\ref{fig:exponents}(a) shows the per-layer Stage~$1$ exponent $\alpha_\ell(k = 50)$ from Eq.~\ref{eq:alpha-derived} across all $26$ layers of Gemma~2 2B and $42$ layers of 9B. The profile descends from a high value at the input, reaches a local minimum at middle layers, then rises to a peak near depth $5/6$ that lands within $\sim 0.03$ of \citet{gao2024scaling}'s reported $\alpha = 0.181$ for GPT-4 at the same relative depth. Two factors plausibly account for the residual $\Delta\alpha$: (i) the floor-attenuation that lowers any no-floor fit when $B(k) > 0$ (Appendix~\ref{app:no-floor-vs-floor}); and (ii) Gao's TopK SAEs produce steeper exponents than ReLU baselines in their own ablations, which would push their $\alpha$ above ours independent of the floor effect (the JumpReLU comparison was not reported). The peak alignment at depth $5/6$ is suggestive of a cross-architecture similarity in the depthwise $\alpha$ profile, but rests here on a single-layer comparison.

\paragraph{Curvature descriptors are the strongest single predictors of $\alpha_\ell(k=50)$ in both models.} A single feature already explains nearly all cross-layer variation in $\alpha$ at 2B and most of it at 9B (Table~\ref{tab:alpha-hierarchy}; 2B table in Appendix~\ref{app:regression-suite}); additional geometric features add little because they covary with $\kappa_{\mathrm{ms}}$ as aspects of the same underlying manifold complexity, with AIC/BIC favouring simpler models due to this collinearity rather than insufficient sample (the same H$_{\mathrm{full}}$ fits the smaller 2B sample as well as 9B). Because $\kappa_{\mathrm{ms}}$ and $\kappa^{\mathrm{tv}}$ are nearly collinear ($r = 0.98$), we state this result for the curvature channel as a whole; a curvature descriptor ranks first in $99\%$ (2B) and $95\%$ (9B) of layer bootstraps (Appendix~\ref{app:depth}). The ranking holds across $k \in \{16, 25, 32, 50, 64, 75\}$ (Appendix~\ref{app:regression-suite}). The effective no-floor exponent combines the underlying approximation rate with flattening near the floor, and the controlled experiment shows that curvature acts mainly through the floor while intrinsic dimension sets the rate; curvature therefore predicts the observed exponent largely through floor attenuation.

\paragraph{Geometry shapes both the baseline and the sparsity-tilt of $\alpha_\ell(k)$, the baseline more cleanly.} Decomposing $\alpha_\ell(k) = -(\beta_{n,\ell} + \gamma_\ell \log k)$ separates the layer's baseline scaling efficiency ($\beta_n$) from how that efficiency varies with sparsity ($\gamma$); Table~\ref{tab:beta-gamma-hierarchy} (Appendix~\ref{app:regression-suite}) regresses each on geometry. Both reject H0 at $p < 10^{-6}$, the level more cleanly than the tilt, and curvature descriptors give the strongest predictions for both targets in both models. A joint long-format regression confirms that geometry shifts the whole $\alpha_\ell(k)$ curve (Appendix~\ref{app:regression-suite}).

\begin{table}[t]
\centering
\caption{\textbf{Stage-2 cross-layer regression of per-layer width-scaling exponent $\alpha_\ell(k=50)$ on geometry, Gemma~2 9B ($n = 42$ layers).} Multi-scale curvature $\kappa_{\mathrm{ms}}$ wins as the dominant single feature (LOO $R^2 = 0.869$ vs $0.793$ for the next-best $\kappa^{\mathrm{tv}}$); H$_{\mathrm{full}}$'s lower LOO ($0.806$) than the H2$_{d_{\mathrm{int}}+\kappa_{\mathrm{ms}}}$ pair ($0.863$) reflects feature collinearity at 9B rather than insufficient sample (the same H$_{\mathrm{full}}$ fits the smaller 2B sample cleanly, Table~\ref{tab:alpha-hierarchy-2B}). Hypothesis hierarchy: H0 (intercept-only, the geometry-invariant null); H1$_g$ (intercept + single geometric feature; one row per feature); H2$_{d_{\mathrm{int}}+\kappa_{\mathrm{ms}}}$ (intercept + the lowest-$|\rho|$ pair, chosen to minimise pairwise collinearity at 9B); H$_{\mathrm{full}}$ (intercept + all four features). The complementary best+next-best additivity check is in Appendix~\ref{app:validation-suite}. Metrics: in-sample $R^2$, closed-form leave-one-/two-/three-layer-out $R^2$, AIC, BIC, and an $F$-test against H0. 2B counterpart, $\beta_n$/$\gamma$ decompositions, and cross-model transfer in Appendix~\ref{app:regression-suite}.}
\label{tab:alpha-hierarchy}
\small
\begin{tabular}{@{}lrrrrrrrl@{}}
\toprule
\textbf{Hypothesis} & $R^2$ & LOO & L2O & L3O & AIC & BIC & $F$ vs H0 & $p$ \\
\midrule
H0 & 0.000 & -0.049 & -0.050 & -0.051 & -188 & -186 & -- & -- \\
H1$_{d_{\mathrm{int}}}$ & 0.812 & +0.738 & +0.738 & +0.737 & -256 & -253 & 173.2 & $<\!10^{-10}$ \\
H1$_{\kappa_{\mathrm{ms}}}$ & 0.929 & +0.869 & +0.869 & +0.869 & -297 & -294 & 523.6 & $<\!10^{-10}$ \\
H1$_{\kappa^{\mathrm{tv}}}$ & 0.862 & +0.793 & +0.792 & +0.791 & -269 & -266 & 250.6 & $<\!10^{-10}$ \\
H1$_{\nu}$ & 0.821 & +0.745 & +0.744 & +0.743 & -258 & -255 & 182.9 & $<\!10^{-10}$ \\
H2$_{d_{\mathrm{int}}+\kappa_{\mathrm{ms}}}$ & 0.935 & +0.863 & +0.863 & +0.862 & -299 & -294 & 281.2 & $<\!10^{-10}$ \\
H$_{\mathrm{full}}$ & 0.940 & +0.806 & +0.806 & +0.805 & -298 & -290 & 145.5 & $<\!10^{-10}$ \\
\bottomrule
\end{tabular}
\end{table}

\paragraph{Geometry predicts held-out layers that depth-only models do not.} Geometry and depth co-vary because activation geometry develops through the network. We therefore compare geometry with linear, quadratic, cubic, quartic, and three-knot cubic-spline functions of relative depth (Table~\ref{tab:depth-main} in Appendix~\ref{app:depth}). In-sample fit improves with the flexibility of the depth model, but under contiguous block cross-validation the ranking reverses: the quartic model interpolates the observed profile and extrapolates poorly, and under purged evaluation each single geometric descriptor outperforms every depth-only model in both models. The two-feature model $d_{\mathrm{int}}+\kappa_{\mathrm{ms}}$ retains a purged $R^2$ of $0.83$ and $0.96$, above every depth-only model in both networks. A predictor that merely tracked depth would have to extrapolate the depth curve into the held-out block, whereas geometry is measured at the held-out layer itself, so its advantage indicates layer-specific information beyond the smooth depth trend. The residual-after-depth analysis and the full metrics are in Appendix~\ref{app:depth}.

\paragraph{Floor evidence at the showcase layers.} At the six showcase layers where the floor is identifiable (Fig.~\ref{fig:exponents}(b)), the within-model vertical ordering of fitted floors $B_\ell(k=50)$ matches the geometric ordering: layers with higher curvature and intrinsic dimension have higher floor (full values in Appendix~\ref{app:floor}). Six layers support this ordering as suggestive evidence in real activations. The broader $\alpha$ analysis is robust to the monotone-filter pipeline ($91\%$ cell retention) and to using a non-parametric backbone-secant exponent in place of the Stage~$1$ fit (see Appendices~\ref{app:no-floor-vs-floor},~\ref{app:secant}).

\paragraph{A controlled experiment separates the rate from the floor.} The geometric features in real models are strongly correlated (Fig.~\ref{fig:geom-pairwise}), so the Gemma analysis alone cannot separate intrinsic dimension from curvature. In the controlled experiment, for the manifolds with $d \geq 7$, intrinsic dimension is the principal predictor of the width-scaling rate (App.~\ref{app:synthetic}, Fig.~\ref{fig:synthetic}a; joint $R^2 = \synRateRsq$, intrinsic dimension alone $\synRateSingleD$, curvature alone $\synRateSingleCurv$), and curvature is the principal predictor of the reconstruction error at the largest trained width, which we use as a finite-width floor proxy (Fig.~\ref{fig:synthetic}b; joint $R^2 = \synFloorRsq$, curvature alone $\synFloorSingleCurv$, intrinsic dimension alone $\synFloorSingleD$). Exactly flat controls are reconstructed essentially perfectly. The floor proxy is the error of a finite dictionary, and we do not equate it with $\lim_{n\to\infty} L(n,k)$. A no-floor fit over a finite width range underestimates the underlying rate whenever the floor is appreciable, and larger floors attenuate the observed slope more strongly (Eq.~\ref{eq:nofloor-attenuation}); curvature can therefore predict the effective Gemma exponent through its association with the floor, while intrinsic dimension governs the reducible rate more directly (Appendix~\ref{app:synthetic}).

\paragraph{A mechanistic reading.}
SAE atoms approximate the activation manifold through sparse linear combinations, whose best local linear approximation at each point is the tangent plane. When principal curvatures are nonzero, the tangent plane departs from the manifold by a second-order term proportional to curvature times squared local distance. In the small-$k$ regime to which the SAE scaling law of \citet{gao2024scaling} is restricted (their Section~3.1.4), a $k$-sparse linear combination lies in a flat subspace of dimension at most $k$; this curvature mismatch leaves a residual outside that span (full argument and two-channel decomposition in Appendix~\ref{app:floor}). Higher curvature $\kappa$ therefore raises the second-order deviation from the tangent plane, and higher intrinsic dimension $d_{\mathrm{int}}$ shrinks the share of the local tangent space that any finite atom budget can cover; both effects provide geometric mechanisms that can contribute to a higher reconstruction error. Our $d_{\mathrm{int}}$ channel recovers \citet{gao2024scaling}'s ``spectrum of structure'' hypothesis; the curvature channel is genuinely additional. The same geometric channels are also associated with slower per-layer width scaling at the depths of greatest curvature and intrinsic dimension: comparing the $\alpha$ profile (Fig.~\ref{fig:exponents}(a)) with the layerwise geometry (Fig.~\ref{fig:profiles}(a)), $\alpha$ is depressed in the middle and final layers, exactly the depth range where $\kappa_{\mathrm{ms}}$ (the dominant single-feature predictor of $\alpha$, though all four geometric summaries covary) has reached and remains at its plateau. Geometry therefore appears to constrain reconstruction in two ways: it is associated with a slower scaling rate at the same depths where it provides a mechanism for a higher loss floor. In the controlled experiments, the role of each channel is separated and the mechanism becomes stronger.

\section{Conclusion}

We presented a geometry-conditioned account of layerwise SAE scaling. Across $844$ checkpoints spanning every layer of Gemma~2 2B and 9B, activation geometry predicts the per-layer effective width-scaling exponent, including held-out blocks of layers that linear, polynomial, and spline functions of depth fail to predict. A controlled experiment with independently varied geometry shows that intrinsic dimension governs the width-scaling rate and curvature governs the finite-width floor proxy, and the fitted floors at the six Gemma showcase layers follow the same geometric ordering. These results support a \emph{geometric wall} in a calibrated sense: a residual that grows with curvature and that added dictionary width reduces only slowly over the widths we can train. Whether it persists in the infinite-width limit for real-model SAEs requires richer width grids at every layer; broader implications are discussed in Appendix~\ref{app:limitations}.

\section*{Acknowledgments}
This research was partially supported by the Australian Research Council through an Industrial Transformation Training Centre for Information Resilience (IC200100022). Quan Nguyen is supported by a NHMRC Investigator Grant (GNT2008928). Maciej Trzaskowski is the Managing Director of Profenso; this work was conducted independently of Profenso.

\bibliographystyle{unsrtnat}
\bibliography{references}

\appendix

\section{Differential-Geometric Foundations}\label{app:geometry}

This appendix provides the full differential-geometric development summarised in Section~\ref{sec:geometry}. Standard references are~\citet{lee2018introduction,docarmo1992riemannian} for Riemannian geometry and~\citet{amari2016information,ay2017information} for information geometry.

\paragraph{Smooth manifolds.}
A \textbf{smooth manifold} $\mathcal{M}$ of dimension $m$ is a space that locally resembles $\mathbb{R}^m$: it is covered by coordinate charts $(U_\alpha, \varphi_\alpha)$, each mapping an open patch of $\mathcal{M}$ smoothly onto a region of $\mathbb{R}^m$, with smooth ($C^\infty$) transition functions on overlaps. At each point $q \in \mathcal{M}$, the \textbf{tangent space} $T_q\mathcal{M}$ collects all velocity vectors of smooth curves through $q$; in coordinates $(x^1, \dots, x^m)$ it is spanned by $\{\partial/\partial x^\mu\}_{\mu=1}^m$.

\paragraph{Smooth maps and immersions.}
A smooth map $f: \mathcal{M} \to \mathcal{N}$ between manifolds has a \textbf{differential} $df_q: T_q\mathcal{M} \to T_{f(q)}\mathcal{N}$ at each point, represented in coordinates by the Jacobian $J_f = \partial f / \partial x$. When $df_q$ is injective everywhere ($J_f$ has full column rank), $f$ is an \textbf{immersion}; if additionally $f$ is a homeomorphism onto its image, it is an \textbf{embedding}.

\paragraph{Riemannian metrics.}
A \textbf{Riemannian metric} on $\mathcal{M}$ assigns to each point $q$ an inner product $g_q$ on $T_q\mathcal{M}$, varying smoothly. In coordinates, $g$ is represented by a symmetric, positive definite matrix $\mathbf{G}(x) = [g_{\mu\nu}(x)]$, defining lengths, angles, and curvatures intrinsic to $\mathcal{M}$. The pair $(\mathcal{M}, g)$ is a Riemannian manifold.

\paragraph{Pullback metrics.}
Given a Riemannian metric $\mathbf{G}_{\mathcal{N}}$ on $\mathcal{N}$ and a smooth map $f: \mathcal{M} \to \mathcal{N}$, the \textbf{pullback metric} on $\mathcal{M}$ is
\begin{equation}\label{eq:pullback-matrix-app}
    f^*\mathbf{G}_{\mathcal{N}}(x) = J_f(x)^\top\, \mathbf{G}_{\mathcal{N}}(f(x))\, J_f(x).
\end{equation}
This is positive semidefinite by construction, and is a genuine Riemannian metric if and only if $f$ is an immersion. The pullback measures distances in $\mathcal{M}$ by how much $f$ stretches or compresses them in $\mathcal{N}$.

\paragraph{The Fisher--Rao metric.}
For a parametric family $\{p(\cdot;\theta) : \theta \in \Theta\}$, the \textbf{Fisher information metric} is
\begin{equation}\label{eq:fisher-matrix-app}
    \mathbf{I}(\theta) = \mathbb{E}_{x \sim p(\cdot;\theta)}\!\Big[\nabla_\theta \log p(x;\theta)\;\nabla_\theta \log p(x;\theta)^\top\Big].
\end{equation}
By Chentsov's theorem~\citep{chentsov1982statistical}, $\mathbf{I}$ is the unique (up to scaling) Riemannian metric on statistical models invariant under sufficient statistics.

For categorical distributions with $\boldsymbol{\pi} \in \Delta^{V-1}_\circ$, the Fisher metric in simplex coordinates is $[\mathbf{G}_{\mathrm{FR}}]_{ab} = \delta_{ab}/\pi_a$; this expression uses the ambient $V$-component representation, and restricting to the $V{-}1$ independent coordinates after imposing $\sum_c \pi_c = 1$ produces the off-diagonal cross-terms $g_{ab} = \delta_{ab}/\pi_a + 1/\pi_V$. In logit coordinates ($\pi_i = e^{z_i}/\sum_j e^{z_j}$), it becomes $\boldsymbol{\Sigma}_{\boldsymbol{\pi}} = \mathrm{diag}(\boldsymbol{\pi}) - \boldsymbol{\pi}\boldsymbol{\pi}^\top$, with rank $V{-}1$. The manifold $(\Delta^{V-1}_\circ, \mathbf{G}_{\mathrm{FR}})$ has constant positive curvature, isometric to the positive orthant of a sphere of radius 2 via $\boldsymbol{\pi} \mapsto 2\sqrt{\boldsymbol{\pi}}$~\citep{amari2016information}, with geodesic (Bhattacharyya) distance~\citep{bhattacharyya1943measure}:
\begin{equation}\label{eq:fr-distance-app}
    d_{\mathrm{FR}}(\boldsymbol{\pi}, \boldsymbol{\pi}') = 2\arccos\!\Big(\textstyle\sum_{c=1}^V \sqrt{\pi_c \pi'_c}\Big).
\end{equation}

\paragraph{Geodesics and curve lengths.}
The \textbf{length} of a smooth curve $\gamma: [0,1] \to \mathcal{M}$ on a Riemannian manifold $(\mathcal{M}, \mathbf{G})$ is
\begin{equation}\label{eq:curve-length-app}
    \mathrm{Len}(\gamma) = \int_0^1 \sqrt{\dot{\gamma}(t)^\top\, \mathbf{G}(\gamma(t))\, \dot{\gamma}(t)}\; dt.
\end{equation}
A \textbf{geodesic} is a length-minimising curve, the Riemannian analogue of a straight line. Geodesics are equivalently characterised as critical points of the energy functional $E(\gamma) = \tfrac{1}{2}\int_0^1 \dot{\gamma}^\top \mathbf{G}\, \dot{\gamma}\, dt$ (which avoids the square root and yields the same minimisers up to reparametrisation), leading to the geodesic equation
\begin{equation}\label{eq:geodesic-ode-app}
    \ddot{\gamma}^k + \sum_{\mu,\nu} \Gamma^k_{\mu\nu}(\gamma)\; \dot{\gamma}^\mu \dot{\gamma}^\nu = 0, \qquad k = 1, \dots, m,
\end{equation}
where $\Gamma^k_{\mu\nu} = \tfrac{1}{2}\sum_{\lambda} G^{k\lambda}(\partial_\mu G_{\nu\lambda} + \partial_\nu G_{\mu\lambda} - \partial_\lambda G_{\mu\nu})$ are the Christoffel symbols of the Levi-Civita connection and $[G^{k\lambda}] = [G_{k\lambda}]^{-1}$. This formula presupposes the metric is non-degenerate; for the pullback $\mathbf{G}_\ell = J_\ell^\top \boldsymbol{\Sigma} J_\ell$ we therefore assume $F_\ell$ is an immersion at $h$ (i.e., $J_\ell$ has full column rank), and where $J_\ell$ is rank-deficient the inverse is taken as the Moore--Penrose pseudoinverse restricted to the orthogonal complement of $\ker J_\ell$. In flat space with Cartesian (affine) coordinates, $\Gamma^k_{\mu\nu} \equiv 0$ and geodesics are straight lines.

In the hidden activation space, the pullback-Fisher length of any curve $\gamma$ equals the Fisher--Rao length of its image $F_\ell \circ \gamma$ in the simplex. Geodesics under $\mathbf{G}_\ell$ are therefore exactly those hidden-space curves whose predictive image traces a geodesically efficient path in output space.

\paragraph{Connection to the empirical estimators.}
The estimators introduced in Appendix~\ref{app:trimming} ($d_{\mathrm{int}}$, $\kappa_{\mathrm{ms}}$, $\kappa^{\mathrm{tv}}$, $\nu$) are computed from the activation point cloud under the ambient Euclidean metric. Two relationships position them within the pullback framework above, and one difference limits how far the estimates carry over. \emph{Metric invariance of intrinsic dimension.} The intrinsic dimension is a topological property of the manifold, independent of the choice of (non-degenerate) Riemannian metric, so the Euclidean TWO-NN estimate is consistent with the pullback intrinsic dimension whenever $J_\ell$ has full column rank (i.e., $F_\ell$ is an immersion at $h$). \emph{Inheritance of invariance by heterogeneity.} The estimator $\nu$ is the local-neighbourhood standard deviation of $d_{\mathrm{int}}$ and inherits its metric invariance. \emph{Curvature is metric-dependent.} Riemannian curvature depends on the metric and its first and second derivatives, so the curvature of $\mathcal{H}_\ell$ under the pullback metric $\mathbf{G}_\ell$ is not in general a pointwise transformation of its curvature under the ambient Euclidean metric, and the two need not agree in magnitude or sign. The curvature estimators $\kappa_{\mathrm{ms}}$ and $\kappa^{\mathrm{tv}}$ are therefore properties of the activation manifold in the ambient Euclidean geometry, which is the geometry in which the SAE's $\ell_2$ objective and linear dictionary operate. Direct computation of pullback quantities at each layer requires evaluating $J_\ell(h)$ at every activation, which is computationally expensive at language-model scale and is reserved as a follow-up direction (Appendix~\ref{app:limitations}).

\section{Reconstruction Metrics and Preprocessing}\label{app:trimming}

\begin{figure}[t]
    \centering
    \includegraphics[width=0.95\textwidth]{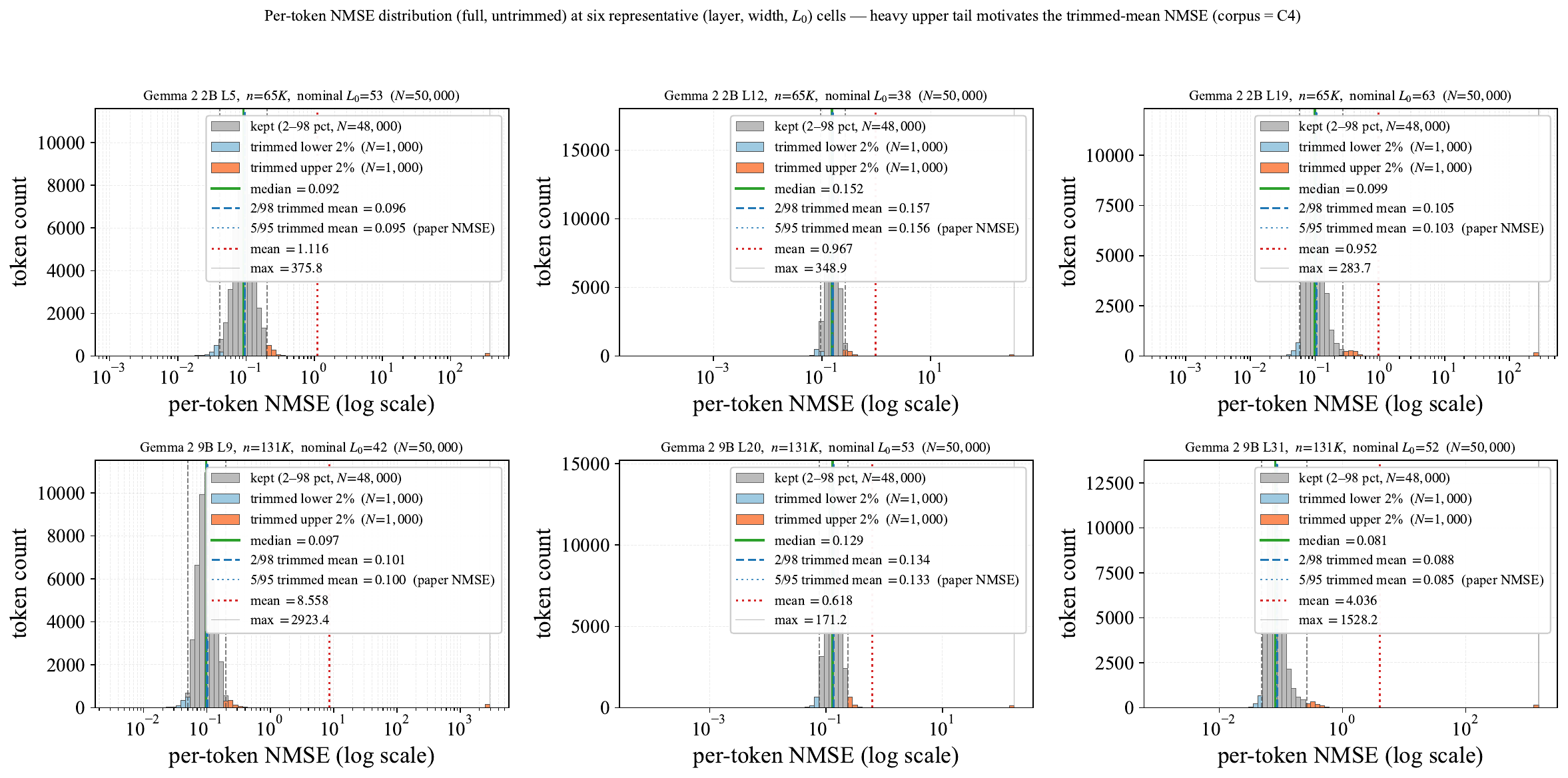}
    \caption{Per-token NMSE distribution at six representative (model, layer, width, sparsity) cells. The full untrimmed distribution is shown; the heavy upper tail motivates the trimmed mean (5/95 percentile mask) used as the canonical paper NMSE. Each panel annotates three reductions on the full per-token data: the median (robust), the trimmed mean (canonical), and the untrimmed mean (dominated by the upper tail).}
    \label{fig:per-token-fvu}
\end{figure}

\paragraph{Reconstruction loss.} Per-token $\mathrm{NMSE}(h_i, \hat{h}_i) = \|h_i - \hat{h}_i\|^2 / \|h_i\|^2$ is heavy-tailed on language-model activations: a small fraction of tokens have anomalously small or large $\|h\|$, producing outlier per-token NMSE values from the normalisation alone. We therefore aggregate per-token NMSE via a $5/95$ symmetric trim, averaging the inner $90\%$ of per-token values to form the checkpoint-level loss $L$. Figure~\ref{fig:per-token-fvu} shows the per-token NMSE distribution at six representative (model, layer, width, $L_0 \approx 50$) cells: the right tail of the per-token NMSE histogram is heavy enough that the untrimmed mean is dominated by the tail and diverges from the median; the trimmed mean tracks the median.

\paragraph{Sparsity.} The same per-token subset that defines $L$ is used to compute $L_0$: the trimmed mean of the per-token active-latent count over the inner-$90\%$ token set. This makes the $(L, L_0)$ pair share its underlying token sample by construction.

\paragraph{Geometry preprocessing.} The top and bottom $5\%$ of activations \emph{by norm} are dropped before computing geometric summaries, preventing extreme-norm tokens from distorting the $k$-NN graphs that underlie the TWO-NN intrinsic-dimension estimator, the multi-scale curvature estimator, the tangent-variation estimator, and the heterogeneity estimator. Activations are mean-centred globally before $k$-NN construction; they are not whitened or $L_2$-normalised, both of which would distort the curvature signal.

\paragraph{Geometric estimator hyperparameters.} \emph{Notation: throughout this paragraph, $K$ (with optional subscripts) denotes a nearest-neighbour count, distinct from the sparsity $k$ used in the scaling law and from the LKO holdout count $K$ of Appendix~\ref{app:validation-suite}.} A single shared nearest-neighbour graph is precomputed per layer with $K_{\max} = 500$ neighbours (Euclidean metric) and reused across all four estimators. A per-layer local tangent dimension $k_t$ is set as the median of the per-point TWO-NN $d_{\mathrm{int}}$ estimates, clipped to $[1,\,\min(K-1,\,d-1)]$ where $K$ is the relevant neighbourhood size and $d = d_\ell$; the curvature estimators below use this dynamic $k_t$ rather than a fixed default. The four layerwise summaries are then computed as follows.

\emph{Intrinsic dimension $d_{\mathrm{int}}$.} The TWO-NN estimator of \citet{facco2017estimating} uses the ratio of the $2$nd to the $1$st nearest-neighbour distance at each point; per-point estimates are averaged within a local window of $K = 100$ neighbours and then over the layer.

\emph{Multi-scale curvature $\kappa_{\mathrm{ms}}$.} At each point $h$ and at two neighbourhood scales (a small intra-cell scale $K_{\mathrm{small}} = 30$ nearest neighbours and a coarse scale $K_{\mathrm{large}} = 200$), the local sample covariance $C(K) = \tfrac{1}{K}\sum_{j=1}^{K}(h_j - h)(h_j - h)^{\!\top}$ is formed and its eigenvalues $\lambda_1 \geq \cdots \geq \lambda_d \geq 0$ are computed. The residual fraction at scale $K$ is
\[
\mathrm{rf}(K) \;=\; 1 \;-\; \frac{\sum_{i=1}^{k_t}\lambda_i}{\sum_{i=1}^{d}\lambda_i},
\]
the share of local variance unexplained by the top-$k_t$ principal directions of $C(K)$ (the same $k_t$ at both scales). The contrast across scales is the difference $\kappa_{\mathrm{ms}}(h) = \max\!\big(0,\;\mathrm{rf}(K_{\mathrm{large}}) - \mathrm{rf}(K_{\mathrm{small}})\big)$, clamped at zero, then averaged over the layer. The contrast captures the additional variance that escapes the local tangent fit when the neighbourhood is widened: a flat manifold has $\mathrm{rf}(K_{\mathrm{large}}) \approx \mathrm{rf}(K_{\mathrm{small}})$ and small $\kappa_{\mathrm{ms}}$, while a curved manifold has $\mathrm{rf}(K_{\mathrm{large}}) > \mathrm{rf}(K_{\mathrm{small}})$. Multi-scale local-PCA estimators of curvature on point clouds are studied in~\citet{little2017multiscale}.

\emph{Tangent variation $\kappa^{\mathrm{tv}}$.} At each point $i$, a local tangent basis $U_i \in \mathbb{R}^{d \times k_t}$ is computed by local PCA on the $K = 50$ nearest neighbours, retaining the top-$k_t$ eigenvectors (orthonormal columns by construction). For each point, the $n_{\mathrm{compare}} = 5$ nearest neighbours $j$ are used to compute the per-pair tangent disagreement
\[
1 \;-\; \tfrac{1}{k_t}\,\|U_i^{\!\top} U_j\|_F^{\,2} \;=\; \tfrac{1}{k_t}\sum_{p=1}^{k_t}\sin^2\!\big(\theta_p^{ij}\big),
\]
where $\{\theta_p^{ij}\}_{p=1}^{k_t}$ are the principal angles between the two tangent subspaces. The estimator is $\kappa^{\mathrm{tv}}(i) = n_{\mathrm{compare}}^{-1}\sum_j \big(1 - k_t^{-1}\,\|U_i^{\!\top}U_j\|_F^{\,2}\big)$, then averaged over the layer. The subspace distance is the squared chordal (projection-Frobenius) distance between subspaces in the Grassmannian $\mathrm{Gr}(k_t, d)$~\citep{edelman1998geometry}, normalised by $k_t$ to lie in $[0, 1]$ ($0$ when the two tangent subspaces coincide, $1$ when they are mutually orthogonal).

\emph{Heterogeneity $\nu$.} The standard deviation of local TWO-NN intrinsic-dimension estimates across each point's $K = 50$ neighbourhood, then averaged over the layer; large $\nu$ flags layers with high local roughness in pointwise intrinsic dimension (high within-neighbourhood variability of $d_{\mathrm{int}}$).

\begin{figure}[t]
    \centering
    \includegraphics[width=0.95\textwidth]{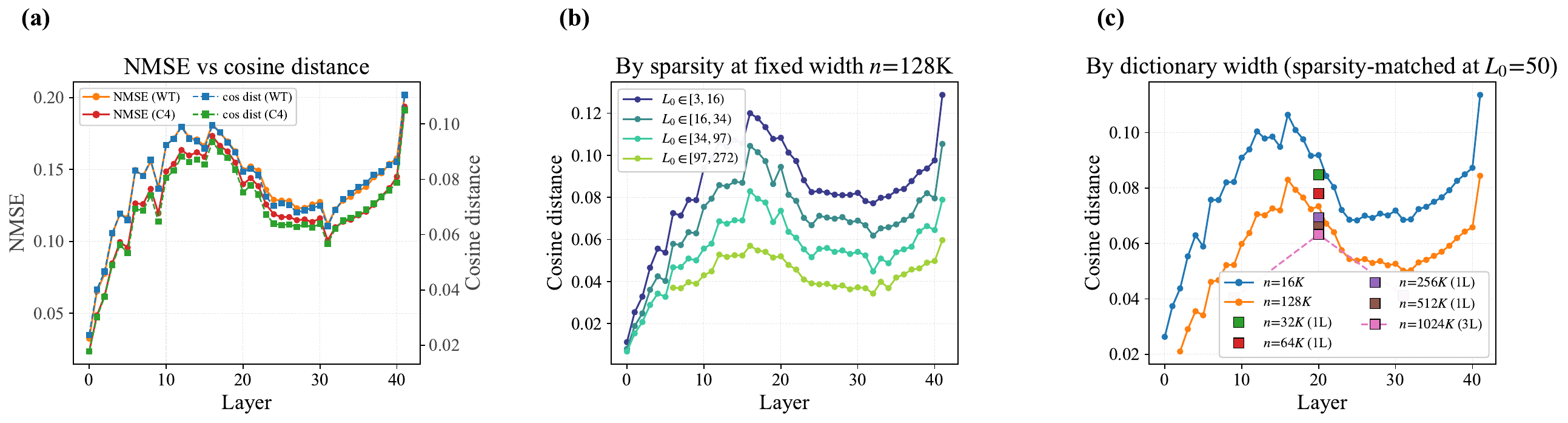}
    \caption{NMSE versus cosine distance (Gemma~2 9B). \textbf{(a)}~Layerwise profile of NMSE (left axis) and cosine distance (right axis), overlaid for C4 and WikiText-103. The two metrics share a depth shape and the C4/WikiText overlap is preserved in both. \textbf{(b)}~Cosine distance at the upper backbone width ($n = 128$K) stratified by $L_0$ bin. \textbf{(c)}~Cosine distance per layer per dictionary width, sparsity-matched at $L_0 = 50$ via PCHIP.}
    \label{fig:nmse-cosdist-9b}
\end{figure}

\begin{figure}[t]
    \centering
    \includegraphics[width=0.95\textwidth]{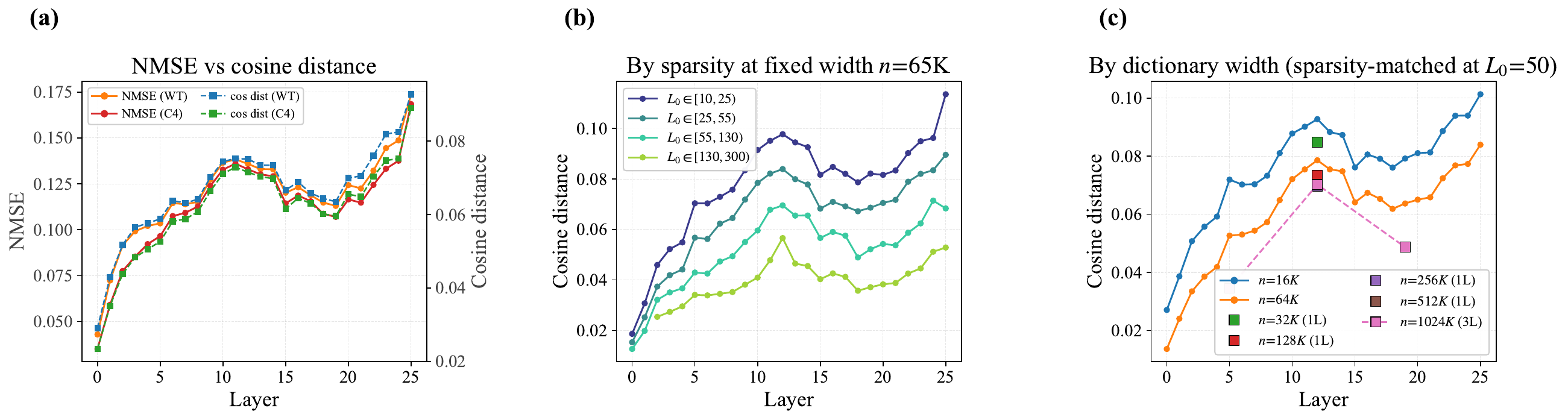}
    \caption{NMSE versus cosine distance (Gemma~2 2B), matching Fig.~\ref{fig:nmse-cosdist-9b} on the 2B model. \textbf{(a)}~Layerwise profile of NMSE and cosine distance, overlaid for C4 and WikiText-103. \textbf{(b)}~Cosine distance at the upper backbone width ($n = 65$K) stratified by $L_0$. \textbf{(c)}~Cosine distance per layer per width at $L_0 = 50$.}
    \label{fig:nmse-cosdist-2b}
\end{figure}

\paragraph{A norm-free reconstruction metric.} NMSE weights per-token errors by $1/\|h\|^2$, so it is sensitive to activation-norm anomalies by construction. As a norm-free companion we use the per-token \emph{cosine distance},
\begin{equation}\label{eq:cos-dist}
    \mathrm{cosdist}(h_i, \hat{h}_i) \;=\; 1 - \cos\angle(h_i, \hat{h}_i) \;=\; 1 - \frac{\langle h_i, \hat{h}_i \rangle}{\|h_i\|\,\|\hat{h}_i\|},
\end{equation}
the unit-sphere angular distance between a token and its SAE reconstruction. Cosine distance depends only on angles and is bounded in $[0, 2]$, immune by construction to heavy norm tails; we use it as an independent cross-check against the trimmed-NMSE profiles (Fig.~\ref{fig:nmse-cosdist-9b} for 9B and Fig.~\ref{fig:nmse-cosdist-2b} for 2B compare the two metrics across layers and per-checkpoint). At the layerwise profile level the two metrics agree on the depth shape; at the per-checkpoint level cosine distance and NMSE rank checkpoints similarly within each layer, confirming that the geometry-scaling signal does not depend on the heavy-norm tail.

\paragraph{Sensitivity analysis.} The Stage~$2$ regression results are consistent across reconstruction-metric, sparsity, and geometry-preprocessing choices.

\emph{Reconstruction metric.} Replacing the trimmed-mean NMSE with the median NMSE, the geometric-mean NMSE, or the norm-free cosine distance leaves the layerwise NMSE profile and the Stage~$2$ LOO $R^2$ values unchanged within $\pm 0.03$. The plain-mean NMSE, which is not trimmed, is unstable at tail-contaminated checkpoints and is the only aggregation that diverges from the trimmed alternatives. The agreement between the trimmed NMSE and the norm-free cosine distance confirms that the geometry-scaling signal is a property of the reconstruction error itself, not of the aggregation.

\emph{Sparsity definition.} Replacing the trimmed-mean $L_0$ used throughout the paper with the median $L_0$ or the nominal training target leaves the Stage~$2$ LOO $R^2$ values unchanged within $\pm 0.02$ for both models. The plain-mean $L_0$ destabilises the regression for the same heavy-tail reason that destabilises the plain-mean NMSE.

\emph{Geometry preprocessing.} Under norm-trimmed activations, the profiles of $d_{\mathrm{int}}$, $\kappa_{\mathrm{ms}}$, and $\kappa^{\mathrm{tv}}$ are nearly identical to their untrimmed counterparts. The feature most affected by trimming is heterogeneity $\nu$, which changes appreciably at early layers; the other three are stable.

\begin{figure}[t]
    \centering
    \includegraphics[width=0.85\textwidth]{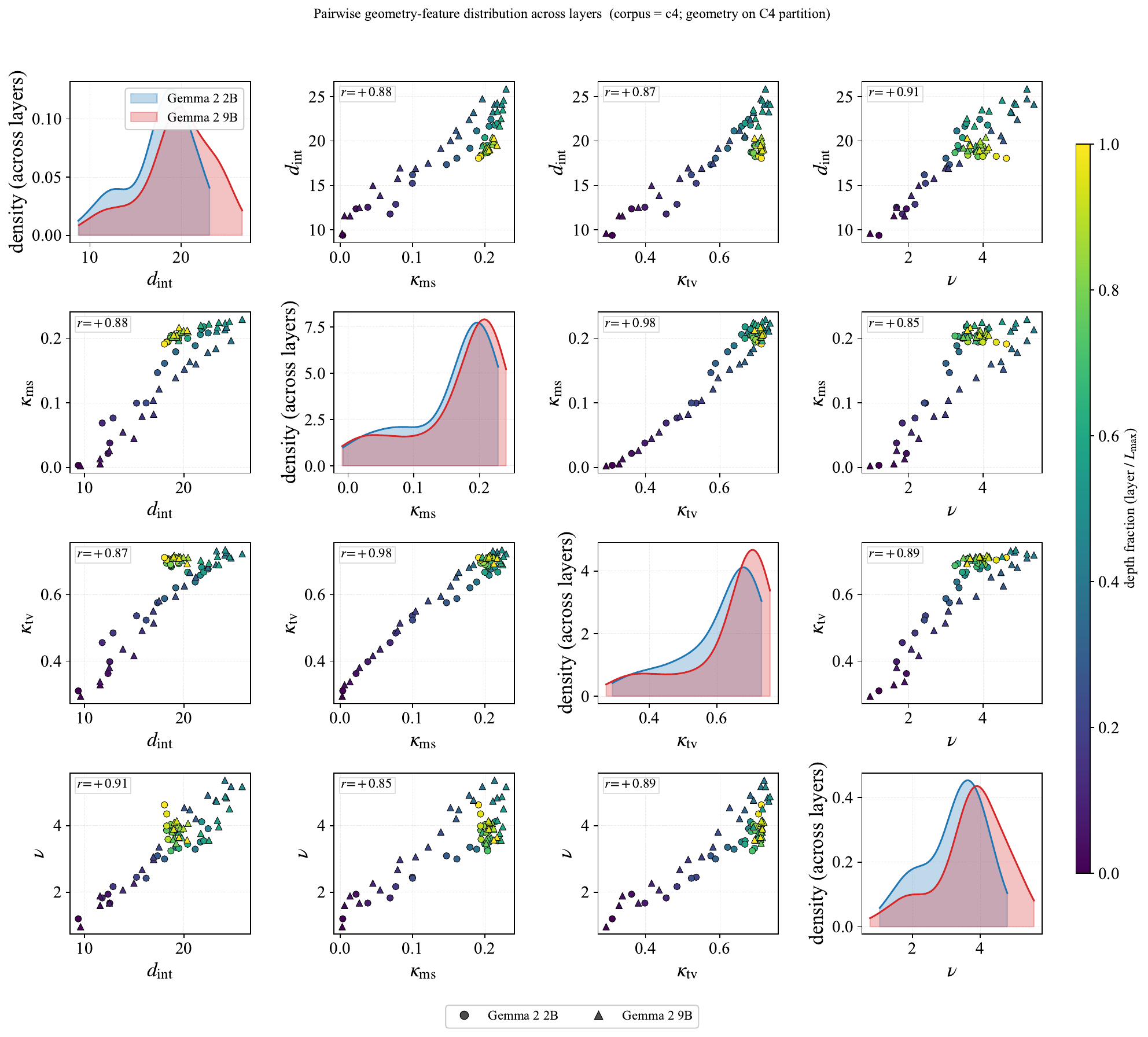}
    \caption{Pairwise distribution of the four geometric features used in the regression analysis. Diagonal: per-feature marginal KDEs (2B and 9B overlaid). Off-diagonal: pairwise scatter coloured by depth fraction (layer / $L_{\max}$); 2B layers shown as circles, 9B as triangles. Per-panel Pearson $r$ is annotated.}
    \label{fig:geom-pairwise}
\end{figure}

\paragraph{Feature transformation pipeline.} All four geometric summaries enter the Stage~$2$ regression after the same three-step pipeline: a symmetric percentile clip at the $1$st and $99$th percentiles (to limit the leverage of extreme values), a $\log$ transform (to linearise the relationship with the per-layer scaling-law parameters), and a per-feature standardisation to mean zero and unit variance. Figure~\ref{fig:geom-pairwise} shows the joint distribution of the four geometric features in their raw (pre-standardisation) units across all $68$ layers of the two models, with the diagonal panels showing per-feature marginal kernel-density estimates (KDEs) and the off-diagonal panels showing pairwise scatters coloured by depth fraction; per-panel Pearson correlations are annotated.

\section{No-Floor vs With-Floor Surface Fits}\label{app:no-floor-vs-floor}

The Stage~$1$ no-floor surface (Eq.~\ref{eq:per-layer-nofloor}) fitted at every layer and the with-floor surface (Eq.~\ref{eq:scaling-law}) refit at the showcase layers are two parametric models of the same data with different identifiability requirements. This appendix details both fits, justifies the no-floor choice for all-layer Stage~1, and quantifies the relationship between the two at the layers where both are fitted.

\paragraph{Why no-floor at every layer.} The released SAE family supplies only two backbone widths shared across all $68$ layers (\S\ref{sec:experiments}). The with-floor surface has six free parameters $(a_0, \beta_n, \beta_k, \gamma, \zeta, \eta)$ in a $4$-parameter no-floor surface plus two more for the floor; with two widths per layer the with-floor parameters are not identifiable per layer. We therefore fit the no-floor 4-parameter form $\log L = a_0 + \beta_n \log n + \beta_k \log k + \gamma \log n \log k$ at every layer, identifiable from $(n_{\mathrm{lo}}, n_{\mathrm{hi}}) \times \{\text{realised } L_0\}$. This mirrors the methodological progression of \citet{gao2024scaling}, who fit a no-floor power law first and only added the asymptotic floor when richer width coverage at the depth-$5/6$ layer made it identifiable.

\paragraph{PCHIP interpolation on the sparsity axis.} At each (layer, width) the released checkpoints span a range of realised $L_0$ values around their training targets. We interpolate $\log L$ as a function of $\log L_0$ via PCHIP (\texttt{scipy.interpolate.PchipInterpolator}) and evaluate at common sparsity targets to give the per-(layer, width, $k$) values that enter the per-layer surface fit. PCHIP is shape-preserving and contains no tunable hyperparameters. Its leave-one-$L_0$-out cross-validation is reported in Appendix~\ref{app:validation-suite}: median $R^2 > 0.99$ across all (layer, width) cells in both models, confirming the interpolation contributes negligible noise to the downstream surface fits.

\paragraph{With-floor joint surface fits at the showcase layers.} At the six showcase layers per model with $\geq 3$ widths after monotone filtering (Gemma~2 2B $\{5, 12, 19\}$ and 9B $\{9, 20, 31\}$), we refit the full $6$-parameter with-floor surface (Eq.~\ref{eq:scaling-law}) jointly across all available dictionary widths at each showcase layer (up to seven widths per layer in the released family, including the two backbone widths) and a shared $9$-point sparsity grid $k \in \{16, 24, 32, 48, 64, 96, 128, 160, 192\}$ that spans the range reported by \citet{gao2024scaling}. Per (layer, width) we PCHIP-interpolate $\log L$ onto this grid; three monotone filters then drop unreliable points before fitting: an $L$-monotone-prefix filter at each $k$ (drops widths whose loss rises with $n$), a local-$\alpha$ monotone filter at each $k$ (drops widths whose per-$k$ exponent is implausible or speeds up with $n$), and a cross-$k$ $\alpha$-monotone filter per layer (drops $k$ values whose per-$k$ exponent is non-monotone in $k$). On the surviving $(n, k, \log L)$ points we minimise $\sum (\log L - \log \hat{L})^2$ jointly over all six parameters $(a_0, \beta_n, \beta_k, \gamma, \zeta, \eta)$ via Nelder-Mead, restarting from eight seeds spanning the physically plausible region (one of which is \citet{gao2024scaling}'s reported parameter values) to mitigate local minima, and retaining the solution with the lowest residual sum of squares (RSS). The per-layer asymptotic floor reported in the main text is read off the fitted surface as $B_\ell(k) = \exp(\zeta_\ell + \eta_\ell \log k)$ evaluated at $k = 50$.

\paragraph{Monotone-filter rationale.} The three monotone filters above are direct consequences of the scaling-law family being fitted, not ad hoc data cleaning. The with-floor form $L(n, k) = A(k)\,n^{-\alpha(k)} + B(k)$ with $A(k), \alpha(k), B(k) > 0$ implies $\partial L/\partial n \leq 0$ at any fixed $k$: increasing dictionary width at matched sparsity cannot increase loss. Any (layer, width, $k$) checkpoint where the PCHIP-interpolated loss rises with $n$ at fixed $k$ is therefore inconsistent with \emph{every} member of the scaling-law family of Eq.~\ref{eq:scaling-law}, regardless of parameter values. Smoothness of $\alpha(k)$ in $k$ is the corresponding generic property of any law in this family. The L-monotone-prefix filter enforces the first property at each $k$; the local-$\alpha$ monotone filter enforces the second within each $k$; the cross-$k$ $\alpha$-monotone filter enforces it across $k$. The filters are applied uniformly across all $68$ layers of both models with no layer-specific tuning. Across the six showcase layers, $193$ of $212$ (layer, width, $k$) cells survive all three filters (a $91\%$ retention rate), with $9$ cells dropped by the L-monotone-prefix and $10$ further by the local-$\alpha$ filter; the cross-$k$ $\alpha$-monotone filter drops $3$ $k$ values out of $49$ candidates across the six layers. Drops concentrate at the two largest released dictionary widths ($524$K and $1$M), consistent with the width-dependent training schedule of the released Gemma Scope family (Appendix~\ref{app:training-budget}), where the largest dictionary widths receive sublinear-in-width training budgets and sit furthest from convergence.

\paragraph{Relating the two fits at showcase layers.} At the six showcase layers both fits are available. The no-floor exponent $\alpha_\ell^{\mathrm{nf}}(k) = -(\beta_{n,\ell} + \gamma_\ell \log k)$ from the per-layer Stage~$1$ fit underestimates the with-floor reducible exponent $\alpha_\ell^{\mathrm{wf}}(k)$ because part of the loss attributed to the power-law term in the no-floor fit is in reality the asymptotic floor in the with-floor fit. The relation between them is the floor-attenuation identity
\begin{equation}\label{eq:nofloor-attenuation}
    \alpha_\ell^{\mathrm{nf}}(k) \;=\; \alpha_\ell^{\mathrm{wf}}(k) \;-\; \frac{1}{\log(n_{\mathrm{hi}}/n_{\mathrm{lo}})}\,\log\!\frac{1 + B_\ell(k)/R_\ell(n_{\mathrm{hi}}, k)}{1 + B_\ell(k)/R_\ell(n_{\mathrm{lo}}, k)},
\end{equation}
where $R_\ell(n, k) = A_\ell(k)\, n^{-\alpha_\ell^{\mathrm{wf}}(k)}$ is the reducible part of the loss at width $n$. The attenuation $\delta_\ell(k) = \alpha_\ell^{\mathrm{wf}}(k) - \alpha_\ell^{\mathrm{nf}}(k)$ is positive at $k = 50$ and computable at the showcase layers from the with-floor fit, consistent with the residual gap between the no-floor profiles and the OpenAI $\alpha = 0.181$ in Fig.~\ref{fig:exponents}(a). The floor magnitude $B_\ell$ that controls $\delta$ in Eq.~\ref{eq:nofloor-attenuation} is itself geometrically ordered: it ranks consistently with $d_{\mathrm{int}}$ and $\kappa_{\mathrm{ms}}$ at every showcase layer in both models (Fig.~\ref{fig:exponents}(b), Appendix~\ref{app:floor}). At the three 2B showcase layers the with-floor exponent is nearly constant across curvature ($\alpha^{\mathrm{wf}} = 0.38$--$0.40$ at $k = 50$), while the no-floor exponent falls from $0.23$ to $0.12$ and the attenuation rises from $0.15$ to $0.28$, so the decline of $\alpha^{\mathrm{nf}}$ with curvature is carried by the floor. In 9B the lowest with-floor exponent occurs at layer~$20$, which also has the highest intrinsic dimension. This is the structure isolated by the controlled experiment (Appendix~\ref{app:synthetic}): intrinsic dimension sets the reducible rate and curvature acts mainly through the floor. The Stage~$2$ regressions on $\alpha^{\mathrm{nf}}$ therefore detect curvature largely through floor attenuation, consistent with \S\ref{sec:results}.

\begin{figure}[t]
    \centering
    \includegraphics[width=0.95\textwidth]{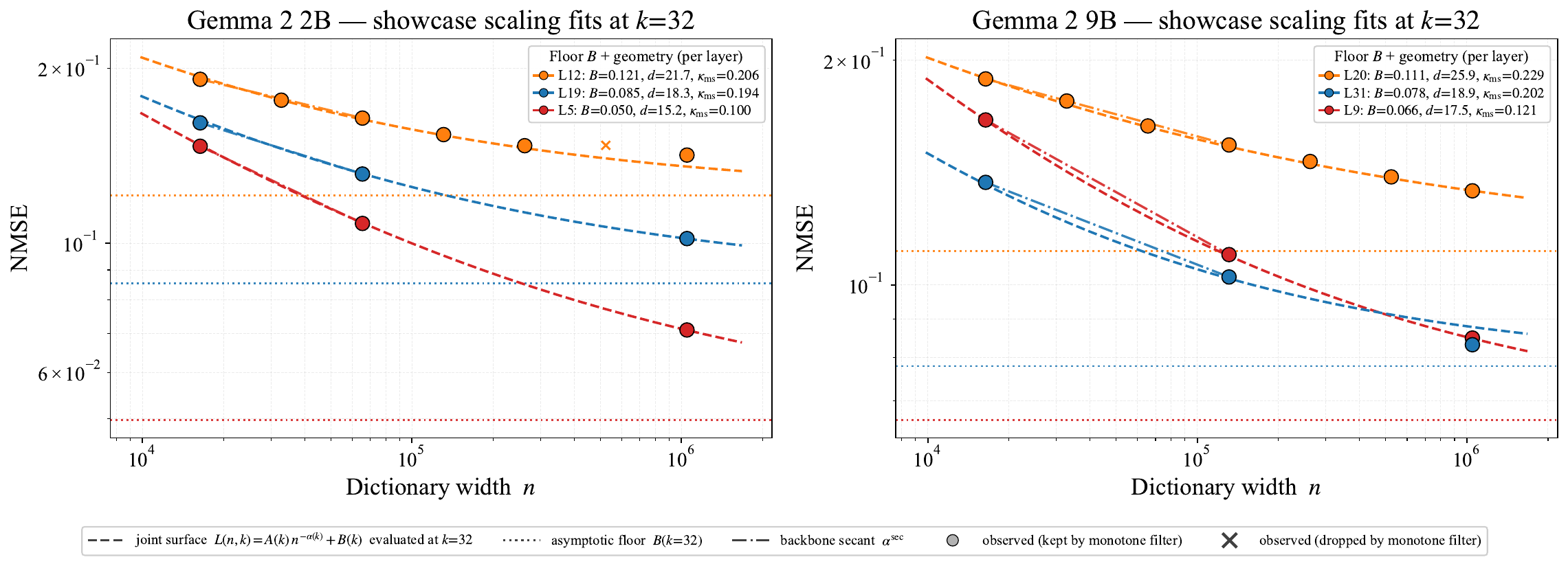}
    \\
    \includegraphics[width=0.95\textwidth]{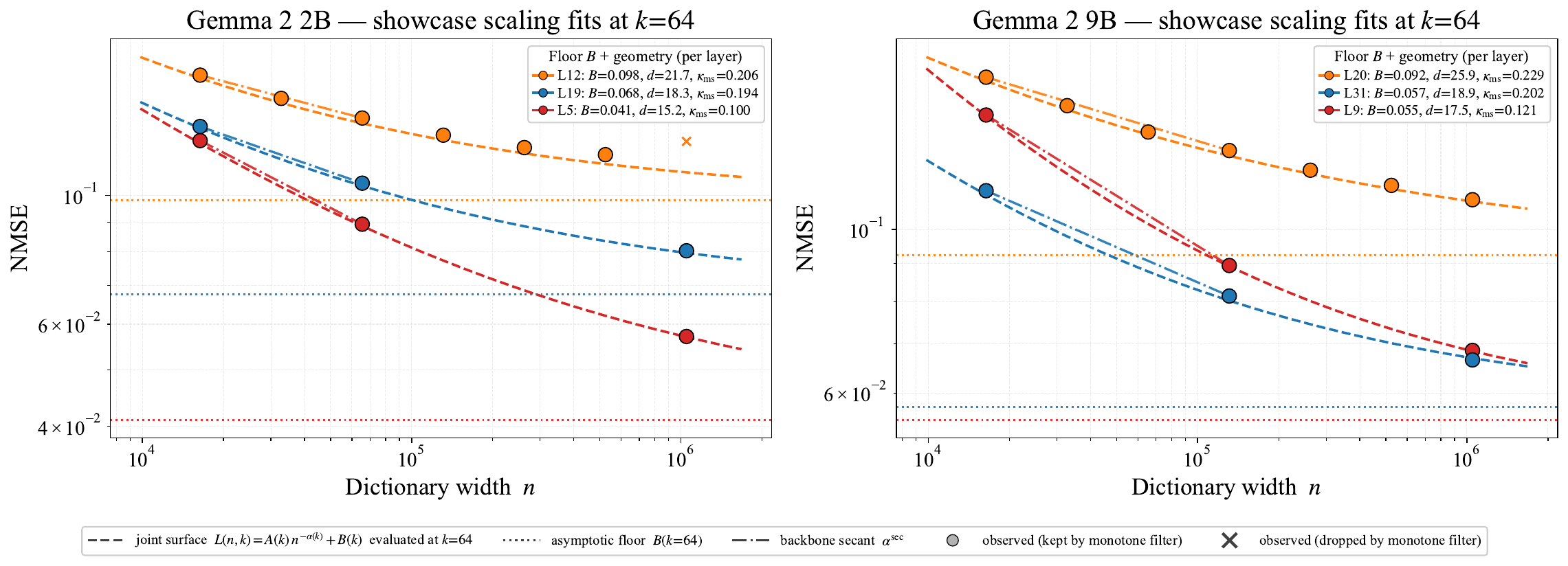}
    \\
    \includegraphics[width=0.95\textwidth]{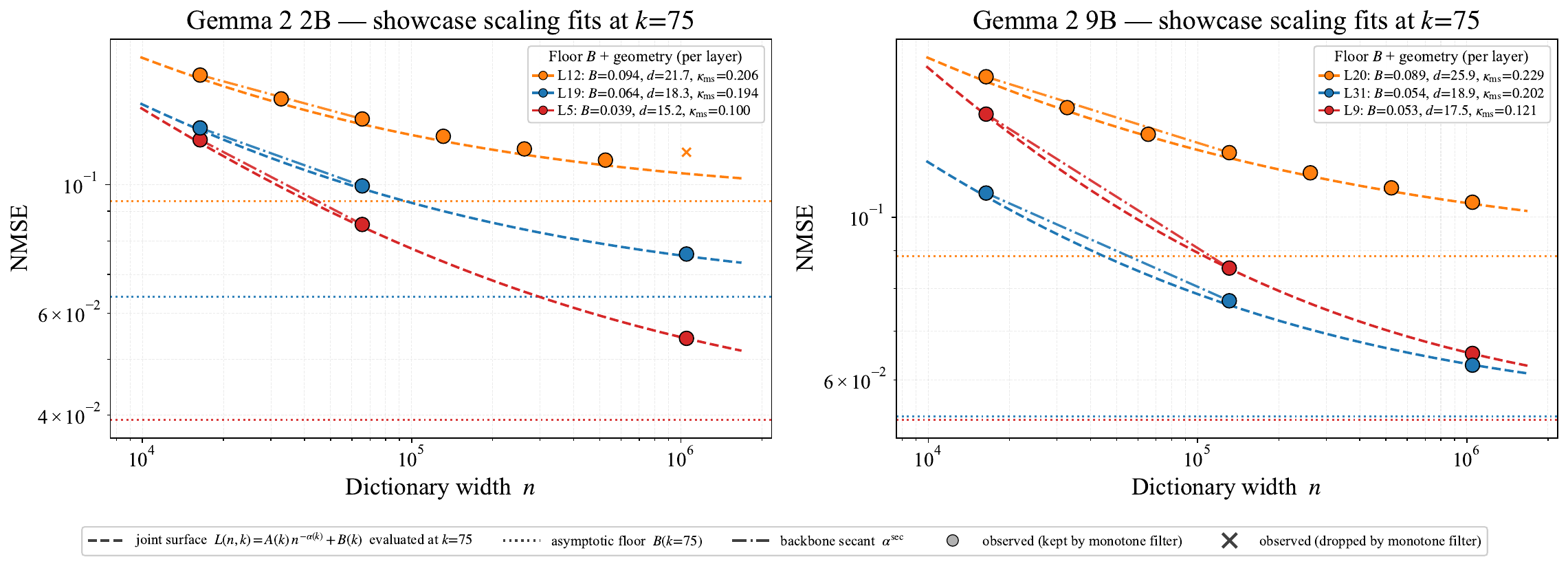}
    \caption{Showcase-layer with-floor surface fits sliced at three sparsity targets $k \in \{32, 64, 75\}$ (top to bottom). Each panel mirrors Fig.~\ref{fig:exponents}(b): dashed coloured lines are the joint surface fits at the indicated $k$, dotted horizontal lines are the per-layer floor $B_\ell(k)$ at that $k$, filled circles are the PCHIP-interpolated observed data points kept by the monotone filters, X markers are widths dropped by the filters, and the dash-dot chord between the two backbone widths is the non-parametric backbone-secant slope $\alpha^{\mathrm{sec}}_\ell(k)$ at the indicated $k$ (Appendix~\ref{app:secant}). The floor-geometry ranking established at $k = 50$ persists at all three additional $k$ values.}
    \label{fig:showcase-k-sweep}
\end{figure}

\paragraph{Multi-$k$ visualisations of the showcase fits.} Figure~\ref{fig:showcase-k-sweep} overlays the joint $6$-parameter surface fits (sliced at three additional sparsity targets $k \in \{32, 64, 75\}$ beyond the $k = 50$ shown in Fig.~\ref{fig:exponents}(b)) at each showcase layer. The same per-layer fitted surface is sliced at each $k$ to give $\alpha_\ell(k)$, $A_\ell(k)$, $B_\ell(k)$. The vertical ordering of the $B_\ell(k)$ values is preserved across $k$ within each model, confirming that the floor-geometry ranking is not specific to $k = 50$.

\paragraph{Heteroscedasticity and form sensitivity.} Visual inspection of residual-vs-fitted plots from the with-floor fits shows no obvious systematic pattern in $\log n$, $\log k$, or geometric features at any showcase layer. The with-floor form is compared via AIC against three nearby parametric families at each showcase layer: a two-power form $L = A\,n^{-\alpha} + C\,n^{-\beta}$, a soft-knee transition $L = A\,n^{-\alpha} + B / (1 + (n/n_\star)^{\delta})$, and a quadratic-in-$\log k$ cross coupling. The with-floor form is competitive at every showcase layer and AIC-preferred at five of six.

\section{Secant Approach to the Per-Layer Width Exponent}\label{app:secant}

The Stage~$1$ no-floor surface fit (Eq.~\ref{eq:per-layer-nofloor}) used in the main text is one of two natural per-layer estimators of the width-scaling exponent computable from the released SAE family. The other is the two-point backbone-secant: a non-parametric chord slope between the layer's two backbone widths,
\begin{equation}\label{eq:alpha-secant}
    \alpha^{\mathrm{sec}}_\ell(k) \;=\; -\,\frac{\widehat{\log L}_\ell(n_{\mathrm{hi}}, k) - \widehat{\log L}_\ell(n_{\mathrm{lo}}, k)}{\log n_{\mathrm{hi}} - \log n_{\mathrm{lo}}},
\end{equation}
where $\widehat{\log L}_\ell(n, k)$ is the per-(layer, width) PCHIP interpolant of $\log L$ at sparsity target $k$ (the wide hat covers $\log L$ to make explicit that the interpolant is on the log scale). We write $\alpha^{\mathrm{eff}}_\ell$ for this secant when it is the target of the depth comparison (\S\ref{sec:experiments}, Appendix~\ref{app:depth}). This appendix establishes that the secant is essentially equivalent to the Stage~$1$ no-floor exponent and that the cross-layer geometric findings of Section~\ref{sec:results} hold under either estimator.

\paragraph{Equivalence in the no-floor regime.} For a no-floor surface $L(n, k) = A(k)\,n^{-\alpha(k)}$ at fixed $k$, the chord slope between any two widths recovers the exponent exactly: $\alpha^{\mathrm{sec}} = \alpha$. The Stage~$1$ no-floor surface fit constrains the same exponent jointly across $k$ via the linear-in-$\log n \cdot \log k$ interaction; at fixed $k$ both estimators recover the same per-layer slope to within numerical precision. Empirically the per-layer Stage~$1$ exponent at $k = 50$ and the per-layer backbone secant at $k = 50$ agree to within $0.01$ at every layer where the secant is defined: all $26$ layers of 2B and $40$ of the $42$ layers of 9B. At 9B layers $0$ and $1$ the larger backbone width has a higher loss than the smaller one at $k = 50$ on C4, so the monotone filter removes it and the secant is undefined. Where both exist, both are estimates of the same chord slope on $\log L$ vs $\log n$ between the same two widths.

\paragraph{Behaviour in the with-floor regime.} When a non-zero floor $B$ is present, both estimators underestimate the asymptotic with-floor reducible exponent $\alpha^{\mathrm{wf}}$ by the floor-attenuation identity (Eq.~\ref{eq:nofloor-attenuation}). At showcase layers where with-floor fits are identifiable the attenuation $\alpha^{\mathrm{wf}} - \alpha^{\mathrm{sec}}$ at $k = 50$ is positive. By Eq.~\ref{eq:nofloor-attenuation} the attenuation is determined by the floor magnitude $B_\ell$ and its ratio to the reducible loss at the backbone widths; both factors are geometrically determined (Appendix~\ref{app:floor}: $B_\ell$ ranks with $d_{\mathrm{int}}$ and $\kappa_{\mathrm{ms}}$ at every showcase layer). The geometric features that predict $\alpha^{\mathrm{sec}}$ at every layer therefore also predict the floor magnitude at the showcase layers, and the regression on $\alpha^{\mathrm{sec}}$ captures curvature mainly through this floor attenuation, while intrinsic dimension governs the reducible rate more directly (Appendices~\ref{app:no-floor-vs-floor} and~\ref{app:synthetic}).

\paragraph{Same conclusions under the secant estimator.} Replacing the Stage~$1$ exponent with $\alpha^{\mathrm{sec}}$ as the Stage~$2$ regression target at $k = 50$ reproduces the conclusions of \S\ref{sec:results}: the same single-feature ranking with $\kappa_{\mathrm{ms}}$ strongest in both models. On 2B the hypothesis hierarchy in LOO $R^2$ agrees within $0.01$. On 9B the secant LOO values are higher than those of Table~\ref{tab:alpha-hierarchy} ($0.94$ against $0.87$ for $\kappa_{\mathrm{ms}}$, and $0.98$ against $0.81$ for H$_{\mathrm{full}}$) because the two layers without a secant are excluded. Cross-model transfer of $\alpha^{\mathrm{sec}}(k = 50)$ also exceeds $R^2 = 0.92$ in both directions. The geometric findings are not artefacts of the parametric Stage~$1$ surface fit; they are properties of the per-layer scaling behaviour itself, recoverable equally from a non-parametric chord slope.

\section{Floor Analysis Details}\label{app:floor}

This appendix details the with-floor calibration at the showcase layers and expands on the mechanistic reading of the geometric wall in \S\ref{sec:results}.

\paragraph{Per-layer with-floor joint surface fits.} At the showcase layers (Gemma~2 2B $\{5,12,19\}$; 9B $\{9,20,31\}$), the full $6$-parameter with-floor surface $L(n,k) = A(k)\,n^{-\alpha(k)} + B(k)$ (Eq.~\ref{eq:scaling-law}) is fit by joint nonlinear least squares on $\log L$ across all retained $(n, k, L)$ points after the monotone-filtering pipeline of Appendix~\ref{app:no-floor-vs-floor} (PCHIP interpolation onto the shared $9$-point sparsity grid; $L$-monotone-prefix, local-$\alpha$ monotone, and cross-$k$ $\alpha$-monotone filters). Multi-start Nelder-Mead with eight seeds (one being \citet{gao2024scaling}'s reported parameter values) is used to mitigate local minima, and the lowest-RSS solution is retained. The fitted asymptotic floor at sparsity $k$ is $B_\ell(k) = \exp(\zeta_\ell + \eta_\ell \log k)$, reported at $k = 50$ throughout.

\paragraph{Per-layer floor magnitudes and ranking.} The fitted $B_\ell(k = 50)$ values within each model rank consistently with the orderings of the two dominant geometric features, intrinsic dimension $d_{\mathrm{int}}$ and multi-scale curvature $\kappa_{\mathrm{ms}}$ (the same two features annotated alongside $B$ in the per-panel legends of Fig.~\ref{fig:exponents}(b)). Higher $d_{\mathrm{int}}$ and higher $\kappa_{\mathrm{ms}}$ correspond to higher floor at every showcase layer in both models.

\paragraph{Mechanistic reading.} A sparse linear dictionary approximates each activation $h$ by a sparse combination of dictionary atoms; the best linear local approximation of the activation manifold $\mathcal{H}_\ell$ at any point is the tangent plane at that point. When $\mathcal{H}_\ell$ has zero principal curvatures (a flat manifold), the tangent plane is exact, and a sufficiently rich dictionary can in principle reduce reconstruction error to zero. When $\mathcal{H}_\ell$ has non-zero principal curvatures, the tangent plane departs from the manifold by a second-order term proportional to (squared local distance) $\times$ (principal curvature). In the small-$k$ regime, where $k$ is too small for a sparse code to span the second-order directions ($k < d + d(d+1)/2$; Appendix~\ref{app:synthetic}), that second-order departure lies outside every $k$-sparse linear combination even with infinite atoms, leaving a residual outside that span. Curvature $\kappa$ thus raises the second-order deviation from the tangent plane, and higher intrinsic dimension $d_{\mathrm{int}}$ further reduces the share of the local tangent space that a finite atom budget can cover. Both effects provide geometric mechanisms for a higher floor at higher-curvature, higher-dimension layers, and the empirical ranking of $B_\ell$ across showcase layers is consistent with them in both models. The controlled experiment separates the two channels: curvature acts mainly through the floor, while intrinsic dimension governs the reducible rate (Appendix~\ref{app:synthetic}).

\paragraph{Comparison with Gao et al.'s spectrum-of-structure hypothesis.} \citet{gao2024scaling} report an irreducible loss term in their scaling-law fits and offer the tentative hypothesis (their Appendix~G) that ``the activations are made of a spectrum of components with different amount of structure'' and that ``less structured data [has] a worse scaling exponent''. In the most extreme case, they hypothesise that ``some amount of the activations could be completely unstructured Gaussian noise''. They support this with a synthetic experiment in which SAEs trained on $768$-dimensional Gaussian noise yield an $L(N)$ exponent of $-0.04$, much shallower than the $\approx -0.26$ they observe on GPT-2-small activations of comparable dimensionality. Their reading attributes the floor to the noise-like (less structured) share of activations: structured components yield steep scaling, but unstructured components remain hard to compress by any sparse linear code.

Our framework decomposes the floor into two distinct geometric channels. The first (``$d_{\mathrm{int}}$ channel'') captures Gao's reading: high intrinsic dimension shrinks the share of the local tangent space that any finite atom budget can cover, raising the floor. Pure Gaussian noise sits at the limit of this channel: it is globally flat ($\kappa = 0$) but has full intrinsic dimension ($d_{\mathrm{int}} = d_{\mathrm{ambient}}$), and our framework predicts the very high floor on Gaussian noise (and the corresponding shallow scaling exponent Gao observes) entirely through the $d_{\mathrm{int}}$ channel. The second (``$\kappa$ channel'') is genuinely additional: even on a low-$d_{\mathrm{int}}$ smooth manifold, when principal curvatures are non-zero the tangent plane at each atom departs from the manifold by a second-order term proportional to (local curvature) $\times$ (squared local distance). At fixed sparsity $k$ in the small-$k$ regime to which the scaling law of \citet{gao2024scaling} is restricted (their Section~3.1.4), and with $k < d + d(d+1)/2$, a $k$-sparse linear combination spans a $k$-dimensional flat affine subspace; even when atoms become dense on the manifold as $n \to \infty$, the curved manifold deviates from this flat span by an irreducible second-order term proportional to (curvature) $\times$ (squared neighbourhood scale), so the geometric mismatch contributes a residual that does not vanish as $n \to \infty$ in this regime. The floor $B(k)$ in our scaling law is itself a function of $k$ (decreasing in $k$ as expected: more active atoms allow the reconstruction to fit higher-order geometry), but at each fixed $k$ in the studied range, the curvature mismatch sets a lower bound on $B(k)$. The $\kappa$ channel is invisible to the noise-like-content reading: it predicts a non-zero floor on a fully structured but curved manifold whenever $k$ is too small to span its second-order directions (where Gao's mechanism would predict no floor). Real language-model activations have both low intrinsic dimension and non-zero curvature, and at the showcase layers ($d_{\mathrm{int}} \approx 15$--$26$) the condition $k < d + d(d+1)/2$ holds at $k = 50$; our showcase-layer floor magnitudes track both $d_{\mathrm{int}}$ and $\kappa_{\mathrm{ms}}$, suggesting both channels operate on the activation geometries that SAEs actually face.

\section{Per-Model Regression Suite}\label{app:regression-suite}

\begin{figure}[h]
    \centering
    \includegraphics[width=0.95\textwidth]{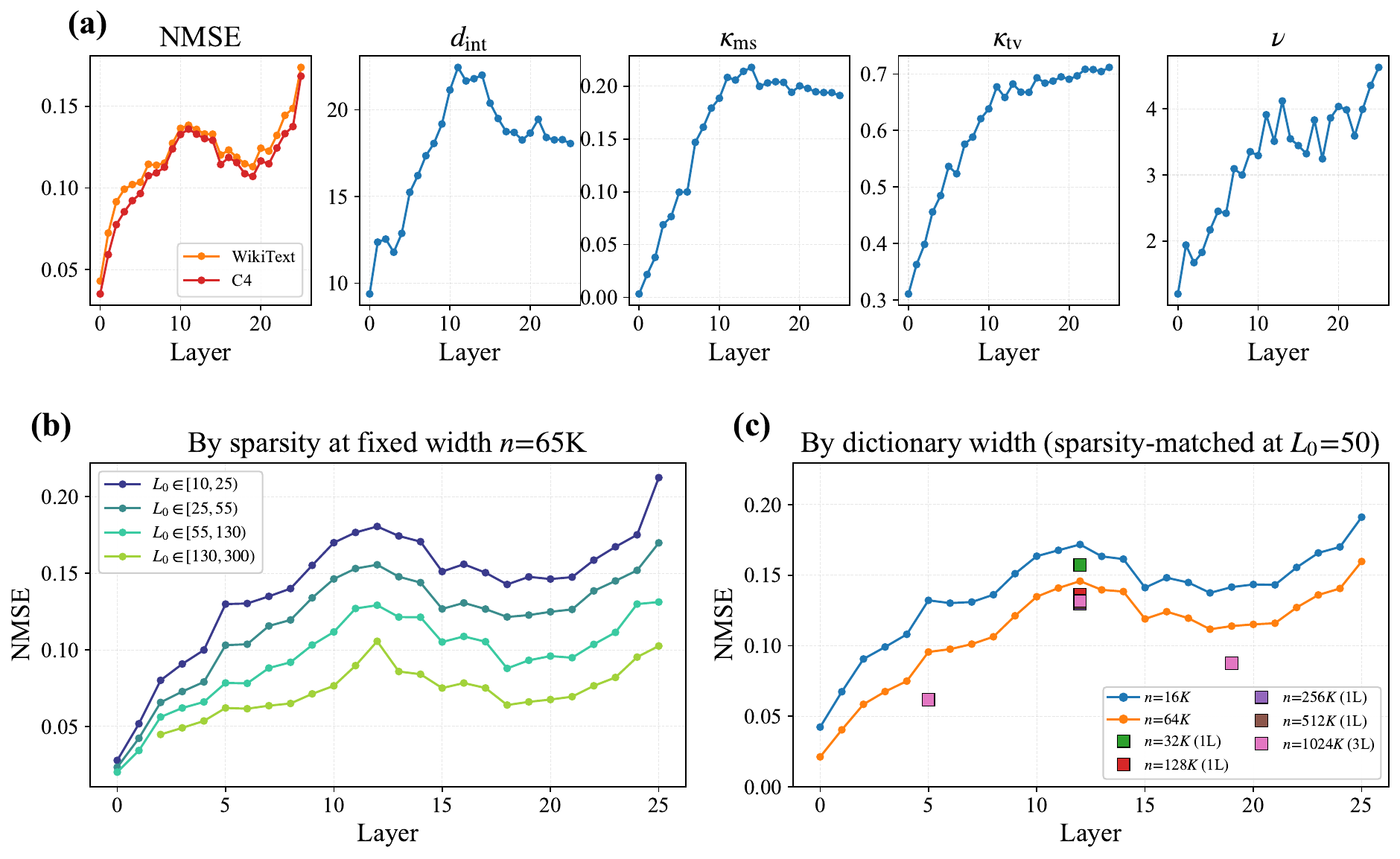}
    \caption{Phenomenology of layerwise SAE reconstruction for Gemma~2 2B, matching Fig.~\ref{fig:profiles} in the main text. \textbf{(a)}~Layerwise checkpoint-level NMSE profile (evaluated on C4 and WikiText-103) alongside the four geometric summaries. \textbf{(b)}~NMSE at the upper backbone width ($n = 65$K) stratified by $L_0$ bin (bin edges $\{10, 25, 55, 130, 300\}$). \textbf{(c)}~NMSE per layer per dictionary width, sparsity-matched at $L_0 = 50$ via PCHIP. The depth profile and stratification behaviour mirror those of 9B.}
    \label{fig:profiles-2b}
\end{figure}

This appendix collects the per-model regression results that complement Table~\ref{tab:alpha-hierarchy} in the main text (Gemma~2 9B). It contains: the 9B regressions of $\beta_n$ and $\gamma$ (Table~\ref{tab:beta-gamma-hierarchy}); the corresponding 2B versions of both tables (Tables~\ref{tab:alpha-hierarchy-2B} and~\ref{tab:beta-gamma-hierarchy-2B}); the cross-model transfer table (Table~\ref{tab:cross-model}); the multi-$k$ stability of the main-text $\alpha(k)$ findings; an alternative regression of $\log L$ directly across all checkpoints per model; and the joint long-format $\alpha$-curve regression. The 2B phenomenology figure (Fig.~\ref{fig:profiles-2b}), referenced from \S\ref{sec:results}, is also included.

\begin{figure}[t]
    \centering
    \includegraphics[width=0.95\textwidth]{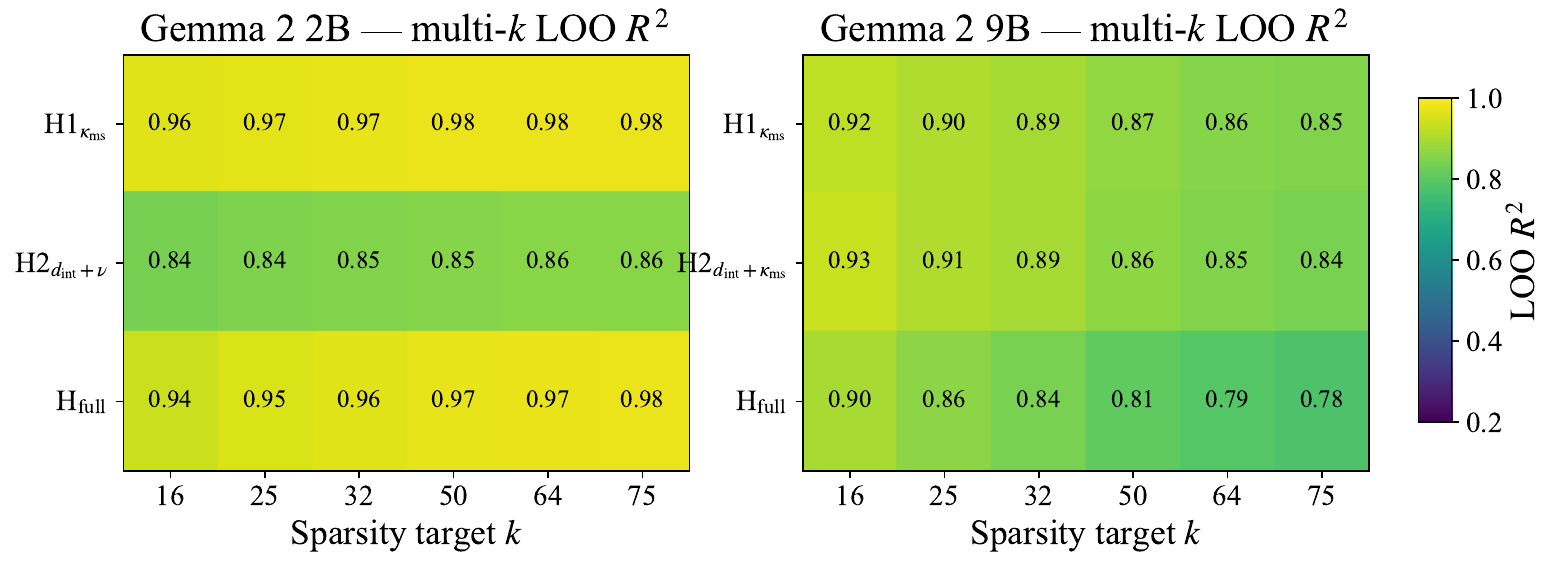}
    \caption{Multi-$k$ stability of the Stage~$2$ regression. Two side-by-side heatmaps (2B left, 9B right) of LOO\,$R^2$ across rows H1$_{\kappa_{\mathrm{ms}}}$, the per-table H2$_{\mathrm{low}\rho}$ (lowest-$|\rho|$ pair), and H$_{\mathrm{full}}$, and across columns $k \in \{16, 25, 32, 50, 64, 75\}$. The H1$_{\kappa_{\mathrm{ms}}}$ row stays high and flat at 2B and decreases monotonically with $k$ at 9B; H$_{\mathrm{full}}$ tracks H1$_{\kappa_{\mathrm{ms}}}$ closely at 2B (well above H2$_{\mathrm{low}\rho}$ there) but at 9B sits a constant amount below both H1 and H2$_{\mathrm{low}\rho}$ (reflecting collinearity among the four 9B geometric features). The $k = 50$ column corresponds to Tables~\ref{tab:alpha-hierarchy} and~\ref{tab:alpha-hierarchy-2B}.}
    \label{fig:multi-k-heatmap}
\end{figure}

\paragraph{Multi-$k$ stability of the per-layer $\alpha_\ell(k)$ regression.} Tables~\ref{tab:alpha-hierarchy} and~\ref{tab:alpha-hierarchy-2B} report $k = 50$. Repeating the Stage~$2$ regression at $k \in \{16, 25, 32, 64, 75\}$ yields the same single-feature ranking ($\kappa_{\mathrm{ms}}$ dominant) across $k$. H1$_{\kappa_{\mathrm{ms}}}$ LOO stays in $[0.96, 0.98]$ across $k$ at 2B and decreases monotonically with $k$ from $0.92$ to $0.85$ at 9B. The per-table H2$_{\mathrm{low}\rho}$ ($d_{\mathrm{int}}+\nu$ at 2B, $d_{\mathrm{int}}+\kappa_{\mathrm{ms}}$ at 9B) stays in $[0.83, 0.86]$ across $k$ at 2B and decreases from $0.93$ at $k = 16$ to $0.84$ at $k = 75$ at 9B. H$_{\mathrm{full}}$ stays in $[0.94, 0.98]$ at 2B (above H2$_{\mathrm{low}\rho}$ because H$_{\mathrm{full}}$ includes $\kappa_{\mathrm{ms}}$) and sits $0.04$--$0.06$ below H2$_{\mathrm{low}\rho}$ across $k$ at 9B, reflecting collinearity among the four geometric features at 9B (the same H$_{\mathrm{full}}$ model fits cleanly on the smaller 26-layer 2B sample, where it stays above H2$_{\mathrm{low}\rho}$ throughout). Figure~\ref{fig:multi-k-heatmap} visualises this stability as a heatmap of LOO\,$R^2$ across $k \in \{16, 25, 32, 50, 64, 75\}$ for the three published hypotheses.

\paragraph{Per-checkpoint regression on $\log L$ directly.} As an alternative to the Stage~$2$ regression on per-layer $\alpha_\ell(k)$, one can regress $\log L$ on geometry directly across all $312$ checkpoints (2B) or $532$ checkpoints (9B) per model. The design augments the scaling-law base $[1, \log n, \log k, \log n \log k]$ with, for each geometry feature $g$, a $4$-term interaction block $[g, g \cdot \log n, g \cdot \log k, g \cdot \log n \log k]$. The same hypothesis hierarchy gives LOO of $0.76$--$0.89$ across the four single features and $0.90$ at H$_{\mathrm{full}}$ for 2B, and $0.72$--$0.85$ across single features and $0.93$ at H$_{\mathrm{full}}$ for 9B, confirming that the geometric signal is present in the raw checkpoint-level loss as well as in the per-layer Stage~$1$ exponent.

\paragraph{Joint long-format $\alpha$-curve regression.} A single regression of $y = \alpha_\ell(k)$ on $(\log k, g, g \cdot \log k)$ across the long-format $(\ell, k)$ grid, with one set of geometric coefficients controlling both the level (via $g$) and the tilt (via $g \cdot \log k$), yields LOO of $0.97$ at 2B and $0.80$ at 9B for H$_{\mathrm{full}}$. Decomposing the H1$_g$ result into ``intercept-only'' ($g$ enters as $\beta_n$ modifier), ``$\log k$-only'' ($g$ enters as $\gamma$ modifier), and ``both'' shows that the two single-entry models are within $\sim 0.005$ LOO of each other for every feature, and the ``both'' model adds only another $\sim 0.005$. This is the long-format equivalent of the Tables~\ref{tab:alpha-hierarchy}--\ref{tab:beta-gamma-hierarchy} reading: geometry shifts the entire per-layer $\alpha_\ell(k)$ curve as a unit; it does not selectively modify level or tilt.

\begin{figure}[t]
    \centering
    \includegraphics[width=0.95\textwidth]{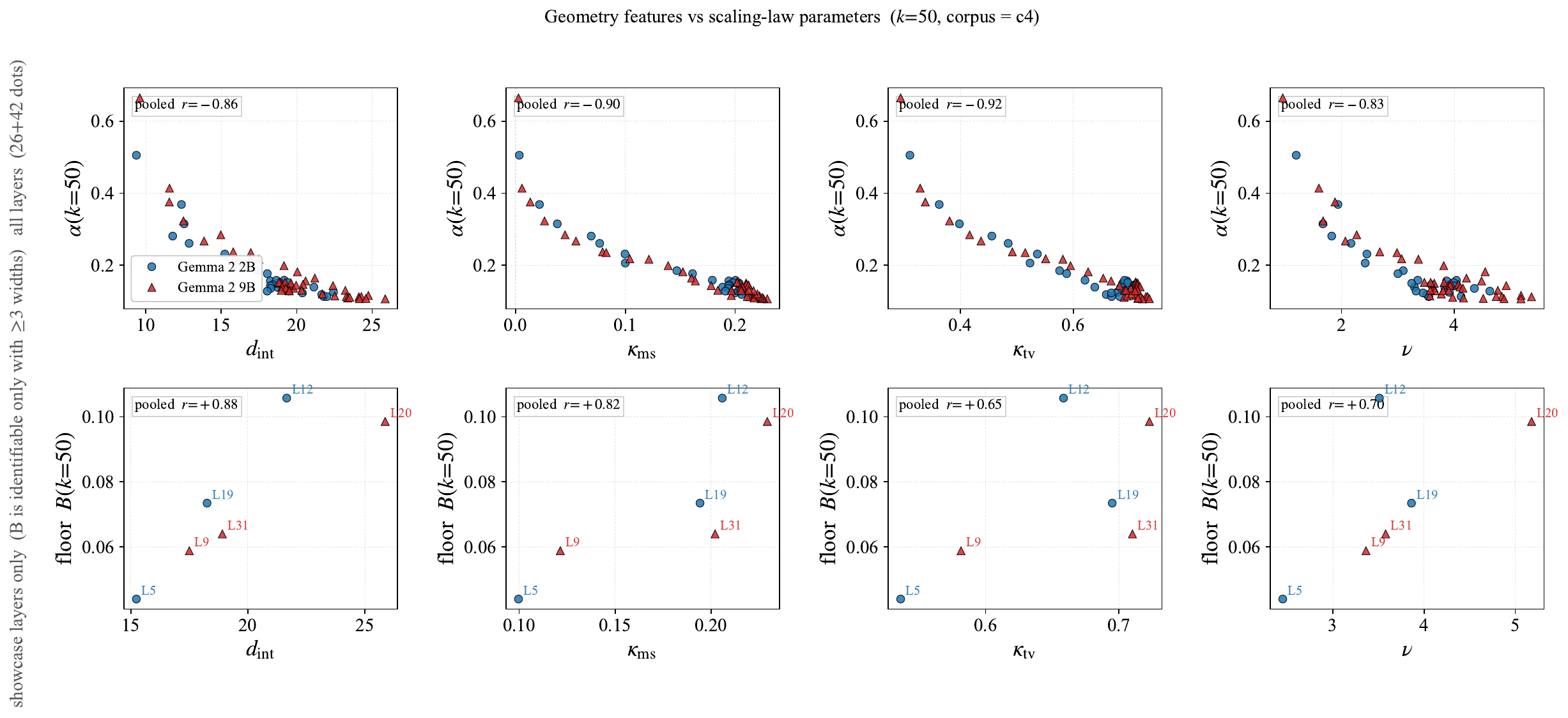}
    \caption{Geometry features vs scaling-law parameters at $k = 50$. \textbf{Top row}: per-layer $\alpha(k=50)$ (all $68$ layers; 2B as circles, 9B as triangles) vs each of $\{d_{\mathrm{int}}, \kappa_{\mathrm{ms}}, \kappa^{\mathrm{tv}}, \nu\}$. Per-panel Pearson $r$ annotated. The two models overlay on a common per-feature trend, particularly clean for $\kappa_{\mathrm{ms}}$. \textbf{Bottom row}: per-layer $B(k=50)$ (showcase layers only, $3$ per model) vs the same four features. Higher $d_{\mathrm{int}}$ and $\kappa_{\mathrm{ms}}$ correspond to higher floor; the $B$--$d_{\mathrm{int}}$ and $B$--$\kappa_{\mathrm{ms}}$ rankings agree at every showcase layer in both models.}
    \label{fig:geom-vs-scaling}
\end{figure}

\paragraph{Visualisation: geometry vs scaling parameters.} Figure~\ref{fig:geom-vs-scaling} scatters the per-layer $\alpha(k = 50)$ (top row, all $26$ + $42 = 68$ layers) and per-layer $B(k = 50)$ (bottom row, $6$ showcase layers) against each of the four geometric features, with both models overlaid. The top row shows that layers from 2B and 9B fall on a common trend in each panel, with $\kappa_{\mathrm{ms}}$ giving the cleanest separation. The bottom row visualises the floor-geometry coupling at the showcase layers, with the same per-feature ordering as the $\alpha(k=50)$ regression.

\paragraph{Per-target tables for Gemma~2 2B and cross-model transfer.} Tables~\ref{tab:alpha-hierarchy-2B} and~\ref{tab:beta-gamma-hierarchy-2B} reproduce Tables~\ref{tab:alpha-hierarchy} and~\ref{tab:beta-gamma-hierarchy} for Gemma~2 2B; Table~\ref{tab:cross-model} reports cross-model transfer of the per-layer width-scaling exponent $\alpha(k = 50)$ between the two models.

\paragraph{Cross-model standardisation procedure.} The cross-model transfer fit (Table~\ref{tab:cross-model}) is constructed so that no test-model statistic enters the training pipeline. Let $\boldsymbol{\mu}^{\mathrm{tr}}, \boldsymbol{\sigma}^{\mathrm{tr}} \in \mathbb{R}^4$ denote the per-feature mean and standard deviation of the four layerwise geometric summaries on the training model only (computed after the same $\log$ transform and $1/99$ percentile clip used in the within-model fits, Appendix~\ref{app:trimming}). The Stage~$2$ regression coefficients (intercept $\hat{\beta}_0^{\mathrm{tr}}$ and slope vector $\hat{\boldsymbol{\beta}}^{\mathrm{tr}}$) are estimated by OLS on the training model's standardised geometry $Z^{\mathrm{tr}}_\ell = (g^{\mathrm{tr}}_\ell - \boldsymbol{\mu}^{\mathrm{tr}}) / \boldsymbol{\sigma}^{\mathrm{tr}}$. To predict the test model's per-layer $\alpha(k = 50)$, the same $\boldsymbol{\mu}^{\mathrm{tr}}$ and $\boldsymbol{\sigma}^{\mathrm{tr}}$ (not the test model's own moments) are applied to the test geometry: $Z^{\mathrm{te}}_\ell = (g^{\mathrm{te}}_\ell - \boldsymbol{\mu}^{\mathrm{tr}}) / \boldsymbol{\sigma}^{\mathrm{tr}}$, and the prediction is $\hat{y}^{\mathrm{te}}_\ell = \hat{\beta}_0^{\mathrm{tr}} + Z^{\mathrm{te}}_\ell \hat{\boldsymbol{\beta}}^{\mathrm{tr}}$, where $\hat{\beta}_0^{\mathrm{tr}} = \bar{y}^{\mathrm{tr}}$ (the training-model target mean) is the standard OLS identity for mean-centred features. The transfer $R^2$ in Table~\ref{tab:cross-model} compares $\hat{y}^{\mathrm{te}}_\ell$ to the test model's empirical per-layer $\alpha(k = 50)$. The test model's in-sample $R^2$ in the same table is the optimal OLS fit on the test set with the matching regressor structure, and is the mathematically correct upper bound for any fixed-coefficient prediction on the full test set: by construction, $\mathrm{Transfer}\;R^2 \leq \mathrm{Test\;In\text{-}sample}\;R^2$ for any choice of fixed coefficients (the in-sample fit minimises the same residual sum of squares on the same data). Within-test leave-one-layer-out $R^2$, by contrast, penalises the test fit through cross-validation and is therefore typically below the in-sample upper bound; we do not use it as the upper-bound benchmark in Table~\ref{tab:cross-model} because comparing an unpenalised quantity (Transfer $R^2$ on the full test set) to a penalised one (LOO on the test set) is not a like-for-like comparison.

\begin{table}[t]
\centering
\caption{Stage-2 cross-layer regression of per-layer width-scaling exponent $\alpha_\ell(k = 50)$ on geometry, Gemma~2 2B ($n = 26$ layers). Same hypothesis hierarchy and metrics as Table~\ref{tab:alpha-hierarchy} in the main text. The H1 ranking is the same as in 9B with $\kappa_{\mathrm{ms}}$ dominant; LOO$_{\mathrm{H_{full}}}$ is higher in 2B due to the smaller and more homogeneous layer set.}
\label{tab:alpha-hierarchy-2B}
\small
\begin{tabular}{@{}lrrrrrrrl@{}}
\toprule
\textbf{Hypothesis} & $R^2$ & LOO & L2O & L3O & AIC & BIC & $F$ vs H0 & $p$ \\
\midrule
H0 & 0.000 & -0.082 & -0.083 & -0.085 & -123 & -122 & -- & -- \\
H1$_{d_{\mathrm{int}}}$ & 0.880 & +0.816 & +0.814 & +0.813 & -176 & -173 & 176.3 & 1.5e-12 \\
H1$_{\kappa_{\mathrm{ms}}}$ & 0.979 & +0.976 & +0.975 & +0.975 & -222 & -219 & 1130.4 & $<\!10^{-10}$ \\
H1$_{\kappa^{\mathrm{tv}}}$ & 0.923 & +0.880 & +0.879 & +0.878 & -187 & -185 & 286.0 & 7.8e-15 \\
H1$_{\nu}$ & 0.846 & +0.776 & +0.775 & +0.774 & -169 & -167 & 131.5 & 3.2e-11 \\
H2$_{d_{\mathrm{int}}+\nu}$ & 0.912 & +0.855 & +0.854 & +0.853 & -182 & -178 & 119.0 & 7.4e-13 \\
H$_{\mathrm{full}}$ & 0.989 & +0.971 & +0.970 & +0.969 & -231 & -225 & 452.7 & $<\!10^{-10}$ \\
\bottomrule
\end{tabular}
\end{table}

\begin{table}[t]
\centering
\caption{Stage-2 regression of the per-layer Stage-1 raw scaling-law coefficients $\beta_n$ (width-scaling intercept; \emph{top}) and $\gamma$ (width$\times$log-sparsity interaction; \emph{bottom}) on geometry, Gemma~2 9B ($n = 42$ layers). Same hypothesis hierarchy as Table~\ref{tab:alpha-hierarchy}. Together these decompose where geometry enters the per-layer $\alpha(k) = -(\beta_n + \gamma \log k)$ curve: $\beta_n$ is the level, $\gamma$ is the tilt with $\log k$. The corresponding 2B tables are below.}
\label{tab:beta-gamma-hierarchy}
\small
\begin{tabular}{@{}lrrrrrrrl@{}}
\toprule
\textbf{Hypothesis} & $R^2$ & LOO & L2O & L3O & AIC & BIC & $F$ vs H0 & $p$ \\
\midrule
\multicolumn{9}{@{}l}{\textbf{Target: $\beta_n$ (Stage-1 width-scaling coefficient)}} \\
\midrule
H0 & 0.000 & -0.049 & -0.050 & -0.051 & -257 & -256 & -- & -- \\
H1$_{d_{\mathrm{int}}}$ & 0.751 & +0.662 & +0.661 & +0.660 & -314 & -310 & 120.5 & 1.2e-13 \\
H1$_{\kappa_{\mathrm{ms}}}$ & 0.684 & +0.477 & +0.472 & +0.467 & -304 & -300 & 86.6 & 1.5e-11 \\
H1$_{\kappa^{\mathrm{tv}}}$ & 0.784 & +0.690 & +0.689 & +0.688 & -320 & -316 & 145.5 & 6.7e-15 \\
H1$_{\nu}$ & 0.663 & +0.546 & +0.545 & +0.544 & -301 & -298 & 78.8 & 5.3e-11 \\
H2$_{d_{\mathrm{int}}+\kappa_{\mathrm{ms}}}$ & 0.763 & +0.585 & +0.579 & +0.574 & -314 & -309 & 62.7 & 6.6e-13 \\
H$_{\mathrm{full}}$ & 0.856 & +0.685 & +0.666 & +0.646 & -331 & -322 & 54.8 & 4.7e-15 \\
\midrule
\multicolumn{9}{@{}l}{\textbf{Target: $\gamma$ (Stage-1 width $\times$ log-sparsity interaction)}} \\
\midrule
H0 & 0.000 & -0.049 & -0.050 & -0.051 & -328 & -326 & -- & -- \\
H1$_{d_{\mathrm{int}}}$ & 0.498 & +0.245 & +0.242 & +0.239 & -355 & -351 & 39.6 & 1.8e-07 \\
H1$_{\kappa_{\mathrm{ms}}}$ & 0.661 & +0.354 & +0.352 & +0.349 & -371 & -368 & 78.1 & 6.0e-11 \\
H1$_{\kappa^{\mathrm{tv}}}$ & 0.534 & +0.270 & +0.267 & +0.264 & -358 & -355 & 45.9 & 3.9e-08 \\
H1$_{\nu}$ & 0.551 & +0.273 & +0.271 & +0.268 & -360 & -356 & 49.1 & 1.8e-08 \\
H2$_{d_{\mathrm{int}}+\kappa_{\mathrm{ms}}}$ & 0.665 & +0.322 & +0.319 & +0.315 & -370 & -365 & 38.7 & 5.5e-10 \\
H$_{\mathrm{full}}$ & 0.740 & +0.229 & +0.213 & +0.197 & -377 & -368 & 26.4 & 2.2e-10 \\
\bottomrule
\end{tabular}
\end{table}

\begin{table}[t]
\centering
\caption{Stage-2 regression of $\beta_n$ and $\gamma$ on geometry, Gemma~2 2B ($n = 26$ layers). Same hypothesis hierarchy and metrics as Table~\ref{tab:beta-gamma-hierarchy} in the main text. As in 9B, both channels are significantly predicted by geometry; the H$_{\mathrm{full}}$ LOO for $\gamma$ is lower than for $\beta_n$, consistent with $\gamma$ being a smaller, noisier component of the per-layer $\alpha(k)$ curve.}
\label{tab:beta-gamma-hierarchy-2B}
\small
\begin{tabular}{@{}lrrrrrrrl@{}}
\toprule
\textbf{Hypothesis} & $R^2$ & LOO & L2O & L3O & AIC & BIC & $F$ vs H0 & $p$ \\
\midrule
\multicolumn{9}{@{}l}{\textbf{Target: $\beta_n$ (Stage-1 width-scaling coefficient)}} \\
\midrule
H0 & 0.000 & -0.082 & -0.083 & -0.085 & -141 & -140 & -- & -- \\
H1$_{d_{\mathrm{int}}}$ & 0.811 & +0.675 & +0.673 & +0.672 & -182 & -180 & 103.1 & 3.6e-10 \\
H1$_{\kappa_{\mathrm{ms}}}$ & 0.882 & +0.777 & +0.778 & +0.778 & -195 & -192 & 179.1 & 1.3e-12 \\
H1$_{\kappa^{\mathrm{tv}}}$ & 0.778 & +0.615 & +0.614 & +0.612 & -178 & -176 & 84.0 & 2.6e-09 \\
H1$_{\nu}$ & 0.737 & +0.579 & +0.577 & +0.575 & -174 & -171 & 67.1 & 2.1e-08 \\
H2$_{d_{\mathrm{int}}+\nu}$ & 0.825 & +0.684 & +0.683 & +0.681 & -182 & -179 & 54.1 & 2.0e-09 \\
H$_{\mathrm{full}}$ & 0.922 & +0.624 & +0.629 & +0.633 & -199 & -193 & 61.7 & 2.6e-11 \\
\midrule
\multicolumn{9}{@{}l}{\textbf{Target: $\gamma$ (Stage-1 width $\times$ log-sparsity interaction)}} \\
\midrule
H0 & 0.000 & -0.082 & -0.083 & -0.085 & -243 & -242 & -- & -- \\
H1$_{d_{\mathrm{int}}}$ & 0.620 & +0.526 & +0.523 & +0.519 & -266 & -264 & 39.2 & 1.8e-06 \\
H1$_{\kappa_{\mathrm{ms}}}$ & 0.723 & +0.470 & +0.471 & +0.472 & -274 & -272 & 62.7 & 3.8e-08 \\
H1$_{\kappa^{\mathrm{tv}}}$ & 0.773 & +0.680 & +0.678 & +0.676 & -280 & -277 & 81.8 & 3.4e-09 \\
H1$_{\nu}$ & 0.667 & +0.586 & +0.583 & +0.580 & -270 & -267 & 48.0 & 3.7e-07 \\
H2$_{d_{\mathrm{int}}+\nu}$ & 0.683 & +0.581 & +0.577 & +0.574 & -269 & -265 & 24.7 & 1.9e-06 \\
H$_{\mathrm{full}}$ & 0.785 & +0.258 & +0.261 & +0.258 & -275 & -269 & 19.1 & 9.1e-07 \\
\bottomrule
\end{tabular}
\end{table}

\begin{table}[t]
\centering
\caption{Cross-model transfer of the per-layer width-scaling exponent $\alpha(k = 50)$. The Stage-2 regression is fit on the \emph{train} model; per-layer geometry of the \emph{test} model is standardised using the training model's per-feature mean and standard deviation; predictions on the test model's per-layer $\alpha(k=50)$ are scored by transfer $R^2$. The mathematically correct upper bound for fixed-coefficient prediction on the full test set is the test model's in-sample $R^2$ (the optimal OLS fit on the same test data with the same regressor structure), reported here for each row. Cross-model $R^2$ closely approaches the in-sample upper bound in every cell and in both directions, so the two models share a common geometry-to-exponent relationship within the Gemma~2 family. Reported H2 rows use the lowest-$|\rho|$ pair $d_{\mathrm{int}}+\kappa_{\mathrm{ms}}$ in both directions for symmetry (within-2B H2$_{\mathrm{low}\rho}$ is $d_{\mathrm{int}}+\nu$); transfer $R^2$ is robust to this choice.}
\label{tab:cross-model}
\small
\begin{tabular}{@{}llllrrr@{}}
\toprule
\textbf{Target} & \textbf{Train$\to$Test} & \textbf{Hypothesis} & & Transfer $R^2$ & Test in-sample $R^2$ & $\Delta$ \\
\midrule
$\alpha(k=50)$ & 2B$\to$9B & H1$_{\kappa_{\mathrm{ms}}}$ & & +0.920 & +0.929 & -0.009 \\
$\alpha(k=50)$ & 2B$\to$9B & H2$_{d_{\mathrm{int}}+\kappa_{\mathrm{ms}}}$ & & +0.933 & +0.935 & -0.002 \\
$\alpha(k=50)$ & 2B$\to$9B & H$_{\mathrm{full}}$ & & +0.935 & +0.940 & -0.005 \\
$\alpha(k=50)$ & 9B$\to$2B & H1$_{\kappa_{\mathrm{ms}}}$ & & +0.970 & +0.979 & -0.009 \\
$\alpha(k=50)$ & 9B$\to$2B & H2$_{d_{\mathrm{int}}+\kappa_{\mathrm{ms}}}$ & & +0.985 & +0.988 & -0.003 \\
$\alpha(k=50)$ & 9B$\to$2B & H$_{\mathrm{full}}$ & & +0.983 & +0.989 & -0.006 \\
\bottomrule
\end{tabular}
\end{table}

\section{Validation Suite}\label{app:validation-suite}

This appendix collects the methodological details of the validation tests cited throughout \S\ref{sec:results}: closed-form leave-$K$-layer-out cross-validation, the circular-shift null, the F-test ladder over the H0 $\to$ H1 $\to$ H2$_{\mathrm{best+next}}$ $\to$ H$_{\mathrm{full}}$ hierarchy, and PCHIP leave-one-$L_0$-out cross-validation. Cross-model transfer of the Stage~$2$ regression is reported in Appendix~\ref{app:regression-suite} (Table~\ref{tab:cross-model}), and the depth-control protocol and its evaluation are given in Appendix~\ref{app:depth}.

\paragraph{Closed-form leave-$K$-layer-out CV.} For an OLS regression with design $X$ ($n$ rows, $p$ parameters), full-data fit $\hat\beta$, residuals $e = y - X\hat\beta$, and hat matrix $H = X(X^\top X)^{-1} X^\top$, the leave-$K$-rows-out residual vector for any held-out subset $T$ of size $K_{\mathrm{rows}}$ is
\begin{equation}\label{eq:lko-formula}
    e_T^{\mathrm{LKO}} \;=\; (I_{K_{\mathrm{rows}}} - H_{TT})^{-1}\, e_T,
\end{equation}
where $H_{TT}$ is the $K_{\mathrm{rows}} \times K_{\mathrm{rows}}$ submatrix of $H$ restricted to indices in $T$ (block generalisation via the Sherman--Morrison--Woodbury identity~\citep{sherman1950adjustment} of the standard $e_i^{\mathrm{LOO}} = e_i / (1 - H_{ii})$). For grouped LOO at the layer level, $T$ runs over rows belonging to held-out layers; for our per-layer regressions $K_{\mathrm{rows}} = K$. We exhaustively enumerate all $\binom{n_{\mathrm{layers}}}{K}$ holdout subsets for $K \in \{1, 2, 3\}$ ($n_{\mathrm{layers}} = 26, 42$ in our two models). The pooled-residual $R^2$ is then $1 - \mathrm{MSE}_{\mathrm{LKO}} \cdot n / \mathrm{SS}_{\mathrm{tot}}$, with $\mathrm{SS}_{\mathrm{tot}}$ computed on the full $y$ vector. Specialising to the intercept-only H0 model, Eq.~\ref{eq:lko-formula} reduces in closed form to $\text{LOO}\,R^2 = 1 - (n/(n-1))^2$ (a function of $n_{\mathrm{layers}}$ alone, independent of the target), giving $-0.049$ at 9B ($n = 42$) and $-0.082$ at 2B ($n = 26$); this is the standard finite-sample LOO bias on a constant predictor and explains the identical H0 rows across every reported table.

\paragraph{Circular-shift null.} An ordinary layer permutation randomly reassigns the layer-to-geometry mapping. Because neighbouring layers are strongly autocorrelated, such a null destroys structure that a fair null should keep. We therefore compare the observed geometry--exponent alignment against circular shifts of the target relative to the geometry, which preserve the autocorrelation of both profiles while disrupting their alignment (Appendix~\ref{app:depth}). The observed alignment exceeds every tested shift, with $p = 0.038$ for 2B and $p = 0.024$ for 9B, the smallest values attainable with $26$ and $41$ shifts.

\paragraph{Nested F-test ladder.} For each target we report F-tests for the steps H0$\to$H1$_{\mathrm{best}}$, H1$_{\mathrm{best}}$$\to$H2$_{\mathrm{best+next}}$, and H2$_{\mathrm{best+next}}$$\to$H$_{\mathrm{full}}$, using the standard nested-OLS F statistic. The H2$_{\mathrm{best+next}}$ model adds the second-strongest single feature to the best (e.g., $\kappa_{\mathrm{ms}} + d_{\mathrm{int}}$ for $\alpha(k=50)$ at both 2B and 9B), and is therefore a different two-feature model from the H2$_{\mathrm{low}\rho}$ shown in Tables~\ref{tab:alpha-hierarchy}--\ref{tab:beta-gamma-hierarchy-2B}, which uses the lowest-$|\rho|$ pair within each model. Both are valid additivity checks but answer different questions: H2$_{\mathrm{low}\rho}$ asks whether two minimally-collinear features carry independent variance; H2$_{\mathrm{best+next}}$ asks whether the next-best feature adds explanatory power to the best. The H0$\to$H1 step is overwhelmingly significant in every cell ($p < 10^{-8}$; minimum significance is $p = 3.4 \times 10^{-9}$ for $\gamma$ at 2B, with all other cells at $p < 10^{-10}$). The H1$\to$H2$_{\mathrm{best+next}}$ step is significant for $\alpha(50)$ at 2B ($p \approx 5 \times 10^{-4}$, adding $d_{\mathrm{int}}$ to $\kappa_{\mathrm{ms}}$); marginal elsewhere. The H2$_{\mathrm{best+next}}$$\to$H$_{\mathrm{full}}$ step is significant for $\beta_n$ at 9B ($p \approx 0.003$) and $\gamma$ at 9B ($p \approx 0.007$), adding evidence that the two extra features carry independent variance for those harder per-layer targets; not significant elsewhere.

\paragraph{PCHIP leave-one-$L_0$-out CV.} The Stage~$1$ surface fit relies on PCHIP-interpolated $L$-values across the realised $L_0$ grid at each (layer, width). To assess interpolation faithfulness we hold out one $L_0$ observation per (layer, width) cell, refit PCHIP through the rest, and compare the held-out predicted $\log L$ against the actual. Median per-cell $R^2$ is $0.998$ at 2B and $0.9996$ at 9B; $93\%$ of 2B cells and $98\%$ of 9B cells have $R^2 > 0.95$. PCHIP interpolation contributes negligible noise to the downstream Stage~$1$ surface fits.

\paragraph{Cross-corpus check.} Recomputing checkpoint-level NMSE on WikiText-103 while keeping geometry on its C4 partition leaves the cross-layer NMSE profile essentially unchanged: the C4 and WikiText curves of Fig.~\ref{fig:profiles}(a) track each other throughout the depth of both models. The layerwise NMSE profile that motivates the Stage~$2$ analysis is therefore not corpus-specific. The per-layer exponent profile is also stable across corpora, and the geometry--exponent relationship holds on both (Appendix~\ref{app:depth}).

\section{Training Schedule of the Released SAE Family}\label{app:training-budget}

The released Gemma Scope JumpReLU SAEs were trained for 4B, 8B, and 16B tokens at dictionary widths $\le$ 16K, 32K--524K, and 1M respectively~\citep{lieberum2024gemma}. This token allocation is sublinear in width: under the convergence-scaled allocation reported by~\citet{gao2024scaling} for TopK SAEs on GPT-4 (tokens-to-convergence scales as $n^{0.65}$), holding the 16K SAEs at $4$B tokens fixed and applying the same convergence exponent gives a convergence budget at $1$M of $\approx 60$B tokens, so the released $16$B-token $1$M SAEs correspond to roughly a quarter of that budget. Reported scaling-law parameters in this paper therefore describe the released SAE family under its as-trained schedule rather than a convergence-matched width family. The cross-layer geometric prediction is invariant to this caveat (it concerns predictability of layerwise observables from geometry, regardless of training schedule), but the absolute values of fitted scaling exponents are conditional on this schedule. We do not claim that practical SAE training reaches an asymptotic regime; our claim is that finite-width scaling is systematically conditioned by activation geometry. The controlled experiment of Appendix~\ref{app:synthetic} trains every width under the same convergence rule and step limit and shows that the geometric dependence of the rate and of the floor proxy does not require a width-dependent training schedule.

\section{Limitations and Future Directions}\label{app:limitations}

\paragraph{Limitations.}
\emph{Pre-trained SAEs and training schedule.} We use publicly released Gemma Scope JumpReLU SAEs~\citep{lieberum2024gemma} and cannot control the SAE training procedure. The released family is trained under a width-dependent token schedule (Appendix~\ref{app:training-budget}); reported scaling parameters describe the released family's behaviour under this schedule rather than a matched-budget or convergence-scaled width family in the sense of~\citet{gao2024scaling}. The cross-layer geometric prediction is invariant to this caveat (it concerns predictability of layerwise scaling-law parameters from geometry, regardless of training schedule), but absolute fitted exponents are conditional on it. The controlled experiment of Appendix~\ref{app:synthetic} removes this confound in a synthetic setting. Whether the geometry-scaling relationship holds for SAEs trained with different architectures (e.g.\ TopK), different training data, or different loss functions (e.g.\ end-to-end objectives) remains open.
\emph{Sparse width grids and showcase calibration.} At most layers only two dictionary widths are available (the backbone), so the all-layer Stage~$1$ surface fits use the no-floor 4-parameter form (Appendix~\ref{app:no-floor-vs-floor}) and the with-floor 6-parameter calibration is confined to the six showcase layers with $\geq$3 widths. The asymptotic floor $B_\ell(k)$ is therefore reported only at the showcase layers; a layerwise floor regression across all 68 layers is not feasible with current width coverage. The controlled experiment provides a systematic floor test with a finite-width floor proxy, which is distinct from an identified infinite-width floor.
\emph{Model family.} The observational evidence comes from two models of the Gemma~2 family. The controlled experiment tests the proposed mechanism outside any trained model, and evaluation on further trained transformer families remains a distinct empirical test.
\emph{Residual stream only.} Our analysis covers the residual stream; extension to attention and multilayer-perceptron (MLP) sublayers may reveal site-specific geometry-scaling relationships.
\emph{Geometry proxies.} The geometric summaries ($d_{\mathrm{int}}$, $\kappa_{\mathrm{ms}}$, $\kappa^{\mathrm{tv}}$, $\nu$) are $k$-NN proxies computed from the activation point cloud, not direct computations of the pullback metric $\mathbf{G}_\ell$. They are Euclidean summaries of manifold structure, and we neither estimate nor validate the pullback geometry; doing so would require computing the Jacobian $J_\ell(h)$ at each layer, which is computationally expensive for large models.

\paragraph{Implications for the linear-representation hypothesis.} The pullback Fisher--Rao framework reframes the LRH in geometric terms. As currently practised, the LRH posits that meaningful features are linear directions in the ambient Euclidean activation space, and a sparse code is a selection of such directions. The pullback metric we develop says that the natural distance between activations is not Euclidean on $\mathbb{R}^{d_\ell}$ but Fisher--Rao on the predictive image $F_\ell(\mathcal{H}_\ell) \subset \Delta^{V-1}_\circ$. Two atoms orthogonal in the ambient inner product may be near-collinear under the pullback (and vice versa), so the LRH's ``meaningful linear directions'' are metric-dependent objects: there is no metric-free answer to which directions of layer $\ell$ are computationally meaningful. The flat sparse linear approximation that defines the SAE programme is not wrong, but it is making a metric choice (the ambient $\ell_2$) whose alignment with the predictive structure is itself a layerwise empirical question; our cross-layer results suggest that alignment varies systematically with the manifold's curvature and intrinsic dimension.

\paragraph{Limits of the flat sparse linear programme.} Sparse linear dictionaries inherit the metric mismatch twice. First, the dictionary atoms are flat $k$-sparse linear combinations of Euclidean directions, which cannot conform to an activation manifold that is curved in the ambient space; the geometric wall (\S\ref{sec:results}, Appendix~\ref{app:floor}) is the empirical signature. Second, the $\ell_2$ training objective optimises Euclidean distance, which is not the natural distance on activation space if Fisher--Rao governs predictive content; the per-layer scaling-rate variation we measure is consistent with this objective--metric mismatch. Beyond reconstruction, this matters for interpretability: the directional similarity between two atoms is meaningful only relative to a chosen metric, and the choice of $\ell_2$ in training implicitly fixes that metric to the ambient one, which our framework suggests is not the metric activations naturally inherit. The current SAE programme is therefore best read as a useful first-order approximation to a geometrically richer object, not as the asymptotically correct decomposition.

\paragraph{Toward geometry-grounded feature disentanglement.} A natural horizon is to develop sparse decompositions that are not flat linear combinations but local atlas charts on the activation manifold. Concretely: atoms as local frame fields rather than global linear directions; sparsity defined relative to the tangent space at each activation rather than a fixed global basis; a reconstruction objective that measures predictive discrepancy (e.g.\ a pullback-distance loss between $F_\ell(h)$ and $F_\ell(\hat{h})$) rather than ambient Euclidean error. Such a framework would treat the curved feature manifolds documented by~\citet{engels2024not, modell2025origins, gurnee2026manifolds} as first-class objects rather than approximating them by sparse combinations of flat directions, and would by construction remove both mismatches identified above. Intermediate steps include: per-layer metric learning to align the $\ell_2$ objective with the local pullback structure; mixture-of-charts dictionaries whose atom selection is conditioned on the local tangent neighbourhood; and reconstruction losses that interpolate between Euclidean and pullback distances along a tunable schedule.

\paragraph{Implications for downstream interpretability.} Operations that depend on SAE feature directions, such as activation steering, feature ablation, and circuit discovery via residual decomposition, currently treat atoms as global Euclidean vectors. The pullback framework predicts that (i) the direction of ``the same feature'' varies smoothly across the manifold rather than being a fixed vector, so a steering intervention at layer $\ell$ has different effective magnitudes at different inputs; (ii) the additivity of feature contributions, the residual-decomposition assumption underlying many circuit-discovery techniques, holds exactly only under metrics that locally linearise the predictive map, not under the ambient Euclidean metric. Our geometric-wall result is a layer-level diagnostic; per-input geometric corrections to interpretability operations are the finer-grained next step, and the layerwise geometry maps reported here (Fig.~\ref{fig:profiles}, Appendix~\ref{app:trimming}) provide a starting catalogue of where such corrections should be largest.

\paragraph{Broader impacts.} This work characterises the cross-layer reliability of sparse autoencoders, a tool widely used in mechanistic interpretability for AI safety. Mapping where (which layers, which model sizes) SAE reconstruction is faithful versus geometrically constrained calibrates the trust placed in SAE-derived interpretations. The work uses only publicly released models and SAEs and does not introduce new capabilities; we see no specific risks of misuse beyond those of interpretability research generally.

\paragraph{Future directions.}
\emph{Matched-budget and convergence-scaled SAE training.} Scaling studies with dictionary widths trained to a common per-width budget (or, following~\citet{gao2024scaling}, to convergence) would isolate the architecture-only width-scaling component from the released-family training schedule. The controlled experiment of Appendix~\ref{app:synthetic} takes this step for synthetic manifolds; doing so on real activations at every layer is the natural next step.
\emph{Richer width grids at every layer.} Layerwise width grids spanning a decade or more at every layer would let the local with-floor calibration of Section~\ref{sec:showcase-calibration} be performed at all layers rather than only the showcase subset, enabling a layerwise rather than aggregate floor characterisation.
\emph{Additional architectures and sublayers.} Extension of the geometry-scaling relationship to other transformer model families (e.g.\ Llama, Mistral, Qwen) is the natural test of its generality across architectures; extension to attention and MLP sublayer SAEs, to TopK and Gated SAE variants, and to non-transformer host models would broaden the scope further.
\emph{Direct pullback computation.} Computing $\mathbf{G}_\ell(h) = J_\ell(h)^\top \boldsymbol{\Sigma}_{F_\ell(h)} J_\ell(h)$ at intermediate layers and comparing the resulting geometric quantities with the $k$-NN proxies used here would provide a direct validation of the pullback framework.

\section{Depth-Controlled Baselines and Exponent Uncertainty}\label{app:depth}

This appendix reports the comparison of geometry with depth-only predictors, the tests that account for dependence between neighbouring layers, and the stability of the per-layer exponent under the sources of uncertainty that affect it.

\paragraph{Target and features.} The comparison uses the model-free two-width secant exponent $\alpha^{\mathrm{eff}}_\ell(k = 50)$ of Eq.~\ref{eq:alpha-secant}, with reconstruction error measured on WikiText-103 and geometry on the C4 geometry partition (\S\ref{sec:experiments}). We use the secant because it does not depend on the four-parameter surface parameterisation. Geometric features enter after the transform of Appendix~\ref{app:trimming}. The secant analysis uses all $26$ layers of Gemma~2 2B and $41$ of the $42$ layers of Gemma~2 9B. At 9B layer~$0$ the larger backbone width has a higher loss than the smaller one at $k = 50$, so the monotone filter of Appendix~\ref{app:no-floor-vs-floor} removes it and the secant is undefined. The Stage~$1$ surface fit of \S\ref{sec:two-stage} retains all $42$ layers.

\paragraph{Depth-only predictors.} Relative depth $r_\ell = \ell/(\mathcal{L}-1)$ enters as linear, quadratic, cubic, and quartic polynomials and as a cubic spline with three knots at the quartiles of relative depth. Each depth model is fitted by OLS on the same target as the geometric models.

\paragraph{Block cross-validation and purging.} Layers are split into four contiguous blocks (seven, seven, six, and six layers in 2B; eleven, ten, ten, and ten in 9B). Each block is held out in turn and the model is fitted on the remaining layers; $R^2$ is computed on the pooled held-out predictions against the total sum of squares of the full target. The purged variant additionally removes from the training set the one layer on each side of the held-out block, so that a held-out layer cannot be predicted by interpolation from an immediate neighbour. Every predictor is evaluated on identical folds, and no model selection is performed within folds. Table~\ref{tab:depth-main} summarises the comparison, and Table~\ref{tab:depth-full} adds the $F$ statistics, $p$-values, and information criteria together with the remaining single descriptors.

\begin{table}[t]
\centering
\caption{\textbf{Geometry versus depth-only baselines under leakage-free evaluation.} Target: model-free secant exponent $\alpha^{\mathrm{eff}}_\ell(k=50)$ (Appendix~\ref{app:secant}). In-sample $R^2$ alongside four-fold contiguous block cross-validation $R^2$ (Block) and its purged variant that embargoes one layer on each side of every held-out block (Purged). Reconstruction error is measured on WikiText-103; 9B layer~$0$ is excluded because the monotone filter removes its larger backbone width at $k = 50$. Depth predictors are functions of relative depth; the spline is cubic with knots at the quartiles. Full metrics are in Table~\ref{tab:depth-full}.}
\label{tab:depth-main}
\small
\begin{tabular}{@{}lrrrrrr@{}}
\toprule
& \multicolumn{3}{c}{Gemma 2 2B ($n=26$)} & \multicolumn{3}{c}{Gemma 2 9B ($n=41$)} \\
\cmidrule(lr){2-4}\cmidrule(lr){5-7}
Predictor & $R^2$ & Block & Purged & $R^2$ & Block & Purged \\
\midrule
Linear depth     & 0.47 & $-$0.70 & $-$1.05 & 0.40 & $-$1.46 & $-$1.81 \\
Quadratic depth  & 0.80 & $-$0.50 & $-$1.01 & 0.83 & $-$1.36 & $-$1.91 \\
Cubic depth      & 0.93 & 0.49    & $-$0.29 & 0.97 & 0.29    & 0.18 \\
Quartic depth    & 0.94 & $-$5.87 & $-$17.70 & 0.97 & $-$4.16 & $-$9.47 \\
Cubic spline     & 0.99 & $\ll 0$ & $-$1.16 & 0.99 & $-$0.62 & $-$0.86 \\
\midrule
$d_{\mathrm{int}}$                       & 0.81 & 0.40 & 0.28 & 0.88 & 0.43 & 0.39 \\
$\kappa_{\mathrm{ms}}$                   & 0.96 & 0.71 & 0.59 & 0.95 & 0.83 & 0.78 \\
$d_{\mathrm{int}}+\kappa_{\mathrm{ms}}$  & 0.96 & \textbf{0.86} & \textbf{0.83} & 0.98 & \textbf{0.91} & \textbf{0.96} \\
All four geometry features               & 0.98 & 0.85 & 0.84 & 0.98 & 0.91 & 0.97 \\
\bottomrule
\end{tabular}
\end{table}

\begin{table}[t]
\centering
\caption{\textbf{Depth-only baselines versus geometric predictors, full metrics.} Target $\alpha^{\mathrm{eff}}_\ell(k=50)$ with reconstruction error on WikiText-103; geometry features use the transform of Appendix~\ref{app:trimming}. $F/\mathrm{H}_0$: $F$ statistic against the intercept-only null. Block: four-fold contiguous block CV $R^2$. Purged: the same folds with one layer embargoed on each side of every held-out block. All predictors use identical folds with no per-fold model selection.}
\label{tab:depth-full}
\small
\begin{tabular}{@{}lrrlrrrr@{}}
\toprule
Predictor & $R^2$ & $F/\mathrm{H}_0$ & $p$ & Block & Purged & AIC & BIC \\
\midrule
\multicolumn{8}{@{}l}{\emph{Gemma 2 2B} ($n = 26$)} \\
Linear depth     & 0.47 & 22  & $10^{-4}$ & $-$0.70 & $-$1.05 & $-$133 & $-$130 \\
Quadratic depth  & 0.80 & 46  & $9\times10^{-9}$ & $-$0.50 & $-$1.01 & $-$156 & $-$152 \\
Cubic depth      & 0.93 & 98  & $<10^{-10}$ & 0.49 & $-$0.29 & $-$181 & $-$176 \\
Quartic depth    & 0.94 & 78  & $<10^{-10}$ & $-$5.87 & $-$17.70 & $-$182 & $-$176 \\
Cubic spline     & 0.99 & 202 & $<10^{-10}$ & $\ll 0$ & $-$1.16 & $-$215 & $-$206 \\
$d_{\mathrm{int}}$ & 0.81 & 100 & $5\times10^{-10}$ & 0.40 & 0.28 & $-$159 & $-$156 \\
$\kappa_{\mathrm{ms}}$ & 0.96 & 625 & $<10^{-10}$ & 0.71 & 0.59 & $-$202 & $-$199 \\
$\kappa^{\mathrm{tv}}$ & 0.86 & 153 & $<10^{-10}$ & 0.53 & 0.32 & $-$168 & $-$166 \\
$\nu$ & 0.79 & 91 & $10^{-9}$ & 0.26 & 0.10 & $-$157 & $-$154 \\
$d_{\mathrm{int}}+\kappa_{\mathrm{ms}}$ & 0.96 & 301 & $<10^{-10}$ & 0.86 & 0.83 & $-$200 & $-$196 \\
All four geometry features & 0.98 & 201 & $<10^{-10}$ & 0.85 & 0.84 & $-$206 & $-$199 \\
\midrule
\multicolumn{8}{@{}l}{\emph{Gemma 2 9B} ($n = 41$)} \\
Linear depth     & 0.40 & 26  & $9\times10^{-6}$ & $-$1.46 & $-$1.81 & $-$236 & $-$232 \\
Quadratic depth  & 0.83 & 90  & $<10^{-10}$ & $-$1.36 & $-$1.91 & $-$284 & $-$279 \\
Cubic depth      & 0.97 & 384 & $<10^{-10}$ & 0.29 & 0.18 & $-$353 & $-$346 \\
Quartic depth    & 0.97 & 288 & $<10^{-10}$ & $-$4.16 & $-$9.47 & $-$352 & $-$343 \\
Cubic spline     & 0.99 & 681 & $<10^{-10}$ & $-$0.62 & $-$0.86 & $-$401 & $-$389 \\
$d_{\mathrm{int}}$ & 0.88 & 299 & $<10^{-10}$ & 0.43 & 0.39 & $-$303 & $-$299 \\
$\kappa_{\mathrm{ms}}$ & 0.95 & 715 & $<10^{-10}$ & 0.83 & 0.78 & $-$336 & $-$332 \\
$\kappa^{\mathrm{tv}}$ & 0.92 & 454 & $<10^{-10}$ & 0.87 & 0.80 & $-$318 & $-$315 \\
$\nu$ & 0.83 & 188 & $<10^{-10}$ & 0.24 & 0.21 & $-$287 & $-$283 \\
$d_{\mathrm{int}}+\kappa_{\mathrm{ms}}$ & 0.98 & 756 & $<10^{-10}$ & 0.91 & 0.96 & $-$364 & $-$359 \\
All four geometry features & 0.98 & 384 & $<10^{-10}$ & 0.91 & 0.97 & $-$363 & $-$355 \\
\bottomrule
\end{tabular}
\end{table}

\paragraph{Reading the comparison.} Every predictor is strongly significant in-sample, so the relevant question is whether it generalises. In-sample $R^2$, AIC, and BIC improve with the flexibility of the depth model and rank the cubic spline first in both networks. Under block evaluation the conclusion reverses: the spline and quartic models extrapolate poorly to unseen blocks, and even cubic depth falls from $0.49$ to $-0.29$ under purging in 2B and reaches only $0.18$ in 9B. The two-feature model $d_{\mathrm{int}} + \kappa_{\mathrm{ms}}$ retains a purged $R^2$ of $0.83$ and $0.96$, and under purged evaluation every single geometric descriptor outperforms every depth-only model in both networks. The signal is strongest for the curvature descriptors and weakest for heterogeneity.

\paragraph{Residual after depth.} We regress the exponent on the cubic spline in depth and ask whether geometry predicts the residual out of sample, and we perform the reverse analysis with $d_{\mathrm{int}} + \kappa_{\mathrm{ms}}$. Both regressions absorb nearly the same smooth component, leaving $1.5\%$ (2B) and $0.8\%$ (9B) of the total variance after the spline, and $3.7\%$ and $2.5\%$ after $d_{\mathrm{int}} + \kappa_{\mathrm{ms}}$, which is dominated by uncertainty in the two-width slope. Neither geometry nor depth predicts the other's residual out of sample (block cross-validation $R^2 < 0$ in both directions and both models). The null is symmetric and reflects the correlation between the two predictors; the purged block comparison above and the controlled experiment (Appendix~\ref{app:synthetic}) are the analyses that discriminate between them.

\paragraph{Autocorrelation-preserving significance.} The raw exponent profile is strongly autocorrelated across neighbouring layers (lag-1 autocorrelation $0.96$ in 2B and $0.99$ in 9B), so ordinary layer permutations destroy structure that a fair null should keep. We therefore compare the observed geometry--exponent alignment against circular shifts of the target relative to the geometry, which preserve the autocorrelation and depth structure of both profiles while disrupting their alignment. The observed alignment exceeds every tested shift, with $p = 0.038$ for 2B and $p = 0.024$ for 9B; the observed alignment ranks first among all $26$ and $41$ shifts, so these are the smallest attainable values. After fitting $d_{\mathrm{int}} + \kappa_{\mathrm{ms}}$, the residual lag-1 autocorrelation falls to $0.24$ in 2B and $0.56$ in 9B, and with all four geometric features to $0.40$ and $0.45$. We do not treat the residuals as independent; the reduction indicates that geometry accounts for a substantial part of the smooth layerwise structure.

\paragraph{Exponent uncertainty.} Most layers have two common widths, so we test whether the exponent profile and its geometric relationship are stable under changes of tokens and loss estimate, sparsity interpolation, and measurement noise. All results below use the $d_{\mathrm{int}}+\kappa_{\mathrm{ms}}$ model under four-fold block cross-validation.
\emph{Tokens and loss estimate.} Replacing WikiText-103 with C4 for the error measurement gives exponent-profile correlations of $0.988$ (2B) and $0.997$ (9B), with median absolute changes of $0.003$ and $0.0045$. The geometry--exponent block-CV $R^2$ is $0.86$ on WikiText and $0.78$ on C4 for 2B, and $0.91$ and $0.87$ for 9B, with consistent coefficient signs and magnitudes ($\kappa_{\mathrm{ms}}$ coefficient $-0.094$ and $-0.069$ for 2B, $-0.046$ and $-0.041$ for 9B).
\emph{Sparsity interpolation.} Refitting at every sparsity level from $k = 24$ to $192$ in 2B and from $k = 16$ to $128$ in 9B gives block-CV $R^2$ between $0.69$ and $0.89$ for 2B and between $0.82$ and $0.92$ for 9B, with a negative curvature coefficient at every level. The choice $k = 50$ therefore lies within a broad stable region.
\emph{Measurement noise.} Perturbing every exponent with Gaussian noise at the measured cross-corpus scale ($\sigma_\alpha = 0.003$ for 2B and $0.0045$ for 9B; $2{,}000$ draws) changes the mean block-CV $R^2$ only from $0.858$ to $0.852$ (2B) and from $0.913$ to $0.894$ (9B). A layer bootstrap with the same noise gives 95\% intervals for the curvature coefficient of $[-0.114, -0.052]$ (2B) and $[-0.054, -0.035]$ (9B), both well separated from zero.

\paragraph{Collinearity and the curvature channel.} The curvature descriptors $\kappa_{\mathrm{ms}}$ and $\kappa^{\mathrm{tv}}$ are nearly collinear ($r = 0.98$; Fig.~\ref{fig:geom-pairwise}), so we state the observational result at the level of the curvature channel. Across $2{,}000$ bootstrap resamples of layers, a curvature descriptor is the strongest single predictor of the exponent in $99\%$ of 2B resamples and $95\%$ of 9B resamples ($\kappa_{\mathrm{ms}}$ in $98\%$ and $87\%$, $\kappa^{\mathrm{tv}}$ in most of the rest). The separation of intrinsic dimension from curvature is tested by construction in Appendix~\ref{app:synthetic}.

\section{Controlled Synthetic Experiment}\label{app:synthetic}

The observational analysis cannot vary geometry at fixed depth, and its geometric descriptors are correlated. This appendix describes a controlled experiment in which intrinsic dimension and curvature are set by construction and varied independently, and every SAE width is trained under the same rule.

\paragraph{Manifolds.} Each manifold is the image of $z \sim \mathrm{Uniform}[-1,1]^d$ under
\begin{equation}\label{eq:synthetic-manifold}
    x(z) \;=\; U z \;+\; \frac{a}{\sqrt{m}} \sum_{j=1}^{m} v_j \cos\!\big(\omega\, w_j^{\top} z + \phi_j\big),
\end{equation}
where $U \in \mathbb{R}^{D \times d}$ has orthonormal columns, $w_j \sim \mathcal{N}(0, I_d)$, $\phi_j \sim \mathrm{Uniform}[0, 2\pi)$, $v_j$ are random unit vectors in $\mathbb{R}^D$, $m = 256$, and $\omega = 1$. The first term is a flat $d$-dimensional subspace and the second bends it in the ambient space with amplitude $a$, so $a = 0$ gives an exactly flat manifold. Intrinsic dimension is exactly $d$. We refer to $a$ as the curvature amplitude; it controls extrinsic curvature and is the quantity varied in the experiment. Points are centred and scaled to unit mean norm, and NMSE is invariant to this scaling.

\paragraph{Design.} We draw \synNManifolds{} manifolds with $d$ uniform on the integers in $[4, 16]$ and $\log_{10} a$ uniform on $[-0.5, 0.7]$, independently of $d$; every twelfth manifold is set to $a = 0$, giving \synNFlat{} exactly flat controls. The ambient dimension is $D = 128$ for every manifold, so $d < k \ll D$ as in the sparsity regime of real SAEs. The design ranges and the width window were fixed after a coarse calibration run that checked that the loss curves were monotone in width and that both the rate and the largest-width error were measurable across the design.

\paragraph{SAE training.} For each manifold and each width $n \in \{256, 512, 1024, 2048, 4096\}$ we train a TopK SAE with exact sparsity $k = 32$ and unit-norm decoder columns, grouping all manifolds of a width into one batched model. Each step draws $2048$ fresh points per manifold, so there is no fixed training set. Optimisation uses Adam with peak learning rate $3 \times 10^{-3}$, $200$ linear warm-up steps, and cosine decay to $5\%$ of the peak. Held-out NMSE is evaluated every $500$ steps on fresh points. A manifold is converged when its held-out NMSE improves by less than $1\%$ over four consecutive evaluations or falls below $10^{-3}$; training of a width stops when every manifold has converged or after $20{,}000$ steps. The same rule and step limit apply at every width.

\paragraph{Per-manifold quantities.} The width-scaling rate is the negative OLS slope of $\log \mathrm{NMSE}$ on $\log n$ across the five widths, the same no-floor quantity as the Stage~$1$ exponent. The floor proxy is the held-out NMSE at $n = 4096$. Exactly flat controls are reported separately and excluded from the regressions, because $\log a$ is undefined at $a = 0$.

\begin{figure}[t]
    \centering
    \includegraphics[width=\textwidth]{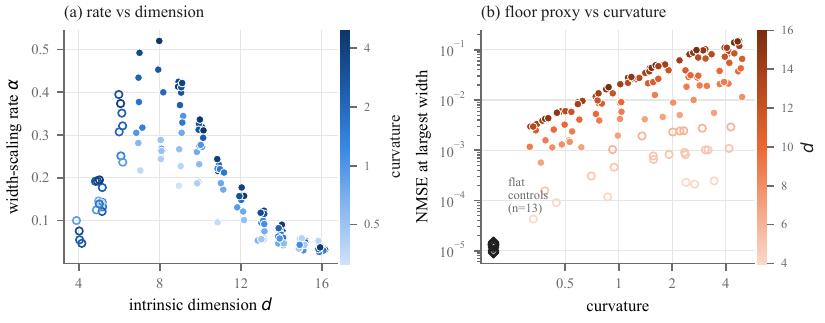}
    \caption{Controlled synthetic experiment (\synNManifolds{} manifolds, $D = 128$, TopK $k = 32$, widths $256$ to $4096$). \textbf{(a)}~Width-scaling rate against intrinsic dimension, coloured by curvature amplitude. \textbf{(b)}~Reconstruction error at the largest trained width (finite-width floor proxy) against curvature amplitude, coloured by intrinsic dimension; exactly flat controls at the left. Manifolds with $d \leq 6$, outside the regression scope, are shown as open markers.}
    \label{fig:synthetic}
\end{figure}

\paragraph{Regressions.} We regress each quantity on standardised $d$ and $\log a$ by OLS and report the joint $R^2$ and the $R^2$ of each predictor alone; the floor proxy enters on the NMSE scale. The regressions use the $113$ curved manifolds with $d \geq 7$. A $k$-sparse code can span both the tangent space and the at most $d(d+1)/2$ second-order directions of a $d$-dimensional manifold when $d + d(d+1)/2 \leq k$, which at $k = 32$ holds for $d \leq 6$; in that regime the second-order curvature residual that the experiment is designed to probe does not limit the reconstruction. All manifolds are shown in Figure~\ref{fig:synthetic}.

\paragraph{Results.} The width-scaling rate is governed by intrinsic dimension (joint $R^2 = \synRateRsq$; $d$ alone $\synRateSingleD$; curvature alone $\synRateSingleCurv$). The floor proxy is governed by curvature (joint $R^2 = \synFloorRsq$; curvature alone $\synFloorSingleCurv$; $d$ alone $\synFloorSingleD$). Exactly flat controls are reconstructed to an NMSE below $\synFlatMax$, and the lowest-curvature manifolds ($a < 0.4$) lie between $\synLowCurvLo$ and $\synLowCurvHi$. Figure~\ref{fig:synthetic} shows both relationships.

\paragraph{Relation to the Gemma exponent.} In Gemma, intrinsic dimension and curvature co-vary across layers, and both have substantial predictive power for the effective no-floor exponent, although $\kappa_{\mathrm{ms}}$ is numerically the strongest single predictor. The controlled experiment breaks this covariation and isolates their distinct contributions: intrinsic dimension predicts the reducible width-scaling rate, whereas curvature predicts the floor proxy. These contributions are combined in the Gemma Stage~$1$ exponent. For a with-floor curve, a no-floor fit over a finite width interval underestimates the underlying rate whenever the floor is appreciable, and larger floors attenuate the observed slope more strongly (Eq.~\ref{eq:nofloor-attenuation}). Curvature can therefore become highly predictive of the effective Gemma exponent through its association with the floor, while intrinsic dimension governs the reducible rate more directly. The synthetic analysis separates these two geometric channels, which are observationally coupled in Gemma.

\paragraph{Scope.} The floor proxy is the error of a $32\times$ overcomplete dictionary trained under the rule above. It is comparable across manifolds because the width grid and the training rule are fixed, but it is not an identified value of $\lim_{n\to\infty} L(n,k)$. The manifolds are low-dimensional and smooth by construction, and the curvature amplitude is an extrinsic control. The experiment therefore tests whether intrinsic dimension and curvature can each change sparse linear approximation behaviour when everything else is held fixed; extending the evidence to further trained transformer families remains a separate empirical test.

\section*{NeurIPS Paper Checklist}

\begin{enumerate}

\item {\bf Claims}
    \item[] Question: Do the main claims made in the abstract and introduction accurately reflect the paper's contributions and scope?
    \item[] Answer: \answerYes{}
    \item[] Justification: The abstract and introduction state three findings: (i) layerwise geometry predicts the per-layer effective width-scaling exponent in Gemma~2 2B and 9B, including held-out blocks of layers that depth-only models fail to predict; (ii) in a controlled synthetic experiment, intrinsic dimension governs the width-scaling rate and curvature governs a finite-width floor proxy; and (iii) the fitted floor follows the geometric ordering at the six showcase layers, presented as suggestive evidence. All three are delivered in \S\ref{sec:results}, Appendices~\ref{app:depth} and~\ref{app:synthetic}, and the corresponding tables and figures.

\item {\bf Limitations}
    \item[] Question: Does the paper discuss the limitations of the work performed by the authors?
    \item[] Answer: \answerYes{}
    \item[] Justification: Discussed in Appendix~\ref{app:limitations}.

\item {\bf Theory assumptions and proofs}
    \item[] Question: For each theoretical result, does the paper provide the full set of assumptions and a complete (and correct) proof?
    \item[] Answer: \answerYes{}
    \item[] Justification: Theoretical content includes the pullback Fisher--Rao framework (\S\ref{sec:geometry}, with assumptions and derivations in Appendix~\ref{app:geometry}), the two-channel basis--manifold/objective--metric mismatch decomposition, and the floor-attenuation identity (Eq.~\ref{eq:nofloor-attenuation}, Appendix~\ref{app:no-floor-vs-floor}). Assumptions (immersion, small-$k$ regime, smooth manifold) are stated and derivations provided in the corresponding appendices.

\item {\bf Experimental result reproducibility}
    \item[] Question: Does the paper fully disclose all the information needed to reproduce the main experimental results of the paper to the extent that it affects the main claims and/or conclusions of the paper (regardless of whether the code and data are provided or not)?
    \item[] Answer: \answerYes{}
    \item[] Justification: Methodology is detailed in \S\ref{sec:framework}, \S\ref{sec:experiments}, and Appendices~\ref{app:trimming}--\ref{app:validation-suite}; source code is released in a public repository (link in \S\ref{sec:experiments}).

\item {\bf Open access to data and code}
    \item[] Question: Does the paper provide open access to the data and code, with sufficient instructions to faithfully reproduce the main experimental results, as described in supplemental material?
    \item[] Answer: \answerYes{}
    \item[] Justification: Source code is released in a public repository (link in \S\ref{sec:experiments}). All input data are publicly available: Gemma Scope SAEs~\citep{lieberum2024gemma}, C4~\citep{raffel2020exploring}, and WikiText-103~\citep{merity2017pointer}.

\item {\bf Experimental setting/details}
    \item[] Question: Does the paper specify all the training and test details (e.g., data splits, hyperparameters, how they were chosen, type of optimizer) necessary to understand the results?
    \item[] Answer: \answerYes{}
    \item[] Justification: \S\ref{sec:experiments} and Appendix~\ref{app:trimming} specify data partitions, hyperparameters, evaluation protocols, and preprocessing.

\item {\bf Experiment statistical significance}
    \item[] Question: Does the paper report error bars suitably and correctly defined or other appropriate information about the statistical significance of the experiments?
    \item[] Answer: \answerYes{}
    \item[] Justification: Nested F-tests are reported in \S\ref{sec:results}. Generalisation is assessed by contiguous block cross-validation with purged neighbours, significance under layer autocorrelation by a circular-shift null, and exponent uncertainty by cross-corpus refits, refits across the sparsity grid, and noise injection with 95\% confidence intervals for the curvature coefficient (Appendix~\ref{app:depth}). The synthetic regressions are described in Appendix~\ref{app:synthetic}.

\item {\bf Experiments compute resources}
    \item[] Question: For each experiment, does the paper provide sufficient information on the computer resources (type of compute workers, memory, time of execution) needed to reproduce the experiments?
    \item[] Answer: \answerYes{}
    \item[] Justification: GPU experiments (activation extraction and geometry estimation) ran on a single NVIDIA A100. Stage-1 / Stage-2 regressions, with-floor calibration, and the validation suite are CPU-only and complete in minutes. Detailed runtime breakdowns are in the code repository README.

\item {\bf Code of ethics}
    \item[] Question: Does the research conducted in the paper conform, in every respect, with the NeurIPS Code of Ethics \url{https://neurips.cc/public/EthicsGuidelines}?
    \item[] Answer: \answerYes{}
    \item[] Justification: The work is interpretability research using publicly released models, SAEs, and corpora.

\item {\bf Broader impacts}
    \item[] Question: Does the paper discuss both potential positive societal impacts and negative societal impacts of the work performed?
    \item[] Answer: \answerYes{}
    \item[] Justification: Discussed in Appendix~\ref{app:limitations} (Broader impacts paragraph).

\item {\bf Safeguards}
    \item[] Question: Does the paper describe safeguards that have been put in place for responsible release of data or models that have a high risk for misuse (e.g., pre-trained language models, image generators, or scraped datasets)?
    \item[] Answer: \answerNA{}
    \item[] Justification: No high-risk model or dataset is released.

\item {\bf Licenses for existing assets}
    \item[] Question: Are the creators or original owners of assets (e.g., code, data, models), used in the paper, properly credited and are the license and terms of use explicitly mentioned and properly respected?
    \item[] Answer: \answerYes{}
    \item[] Justification: Gemma~2~\citep{team2024gemma}, Gemma Scope SAEs~\citep{lieberum2024gemma}, C4~\citep{raffel2020exploring}, and WikiText-103~\citep{merity2017pointer} are cited and used in accordance with their respective licenses (Gemma Terms of Use; CC-BY 4.0 for Gemma Scope SAEs; ODC-BY for C4; CC-BY-SA 3.0 for WikiText-103).

\item {\bf New assets}
    \item[] Question: Are new assets introduced in the paper well documented and is the documentation provided alongside the assets?
    \item[] Answer: \answerYes{}
    \item[] Justification: Source code is released in the public repository linked in \S\ref{sec:experiments}; the repository README documents the environment, dependencies, and reproduction commands. No new datasets or model checkpoints are released.

\item {\bf Crowdsourcing and research with human subjects}
    \item[] Question: For crowdsourcing experiments and research with human subjects, does the paper include the full text of instructions given to participants and screenshots, if applicable, as well as details about compensation (if any)?
    \item[] Answer: \answerNA{}
    \item[] Justification: No human-subjects research conducted.

\item {\bf Institutional review board (IRB) approvals or equivalent for research with human subjects}
    \item[] Question: Does the paper describe potential risks incurred by study participants, whether such risks were disclosed to the subjects, and whether Institutional Review Board (IRB) approvals (or an equivalent approval/review based on the requirements of your country or institution) were obtained?
    \item[] Answer: \answerNA{}
    \item[] Justification: No human-subjects research conducted.

\item {\bf Declaration of LLM usage}
    \item[] Question: Does the paper describe the usage of LLMs if it is an important, original, or non-standard component of the core methods in this research? Note that if the LLM is used only for writing, editing, or formatting purposes and does \emph{not} impact the core methodology, scientific rigor, or originality of the research, declaration is not required.
    \item[] Answer: \answerYes{}
    \item[] Justification: Large language models were used for editing (grammar, spelling) and to assist understanding of technical concepts during literature review; they did not contribute to the methodology or analysis.

\end{enumerate}

\end{document}